\documentclass[10pt,twocolumn,letterpaper]{article}

\usepackage[pagenumbers]{cvpr} 

\usepackage[dvipsnames]{xcolor}

\newcommand{\figref}[1]{Fig.~\ref{#1}}%
\newcommand{\tabref}[1]{Tab.~\ref{#1}}%
\newcommand{\secref}[1]{Sec.~\ref{#1}}
\renewcommand{\eqref}[1]{Eq.~(\ref{#1})}

\usepackage{times}
\usepackage{graphicx}
\usepackage{overpic}
\usepackage{amsmath}
\usepackage{amssymb}
\usepackage{amsmath}
\usepackage{algorithm}
\usepackage{listings}

\usepackage{etoolbox}
\makeatletter
\AfterEndEnvironment{algorithm}{\let\@algcomment\relax}
\AtEndEnvironment{algorithm}{\kern2pt\hrule\relax\vskip3pt\@algcomment}
\let\@algcomment\relax
\newcommand\algcomment[1]{\def\@algcomment{\footnotesize#1}}
\renewcommand\fs@ruled{\def\@fs@cfont{\bfseries}\let\@fs@capt\floatc@ruled
  \def\@fs@pre{\hrule height.8pt depth0pt \kern2pt}%
  \def\@fs@post{}%
  \def\@fs@mid{\kern2pt\hrule\kern2pt}%
  \let\@fs@iftopcapt\iftrue}
\makeatother

\usepackage{makecell}

\definecolor{cvprblue}{rgb}{0.21,0.49,0.74}
\usepackage{color}
\usepackage{xcolor}

\usepackage[pagebackref,breaklinks,colorlinks,citecolor=cvprblue]{hyperref}

\usepackage{overpic}
\usepackage{booktabs}
\usepackage{multirow}
\usepackage{pifont}
\usepackage{enumitem}
\usepackage{bm}
\usepackage{diagbox}

\graphicspath{{./figs/}}
\DeclareGraphicsExtensions{.pdf,.jpg,.png}

\newcommand{\ourMthd}{StyleExpert}
\newcommand{\myPara}[1]{\vspace{2pt}\noindent\textbf{#1}}
\usepackage[normalem]{ulem}
\useunder{\uline}{\ul}{}

\def\paperID{33442} 
\def\confName{CVPR}
\def\confYear{2026}
\definecolor{lightgreen}{RGB}{64,128,128}

\title{Mixture of Style Experts for Diverse Image Stylization}

\author{
    \makebox[\textwidth]
    {\normalsize
        Shihao Zhu$^{1}$ \quad Ziheng Ouyang$^{1}$ \quad Yijia Kang$^{1}$ \quad Qilong Wang$^{3}$ \quad Mi Zhou$^{4}$ \quad Bo Li$^{4}$ 
    } \\
    \makebox[\textwidth]
    {\normalsize 
     Ming-Ming Cheng$^{1,2}$ \quad Qibin Hou$^{1,2}$\footnotemark
    } \\
    \makebox[\textwidth]
    {\normalsize
    $^1$VCIP, CS, Nankai University \quad  $^2$NKIARI, Shenzhen Futian \quad   $^3$Tianjin University \quad $^4$vivo BlueImage Lab
    } \\
}

\begin{document}

\twocolumn[{%
  \maketitle
  \centering
  \captionsetup{type=figure}
  \vspace{-0.7cm}    
  \includegraphics[width=\linewidth]{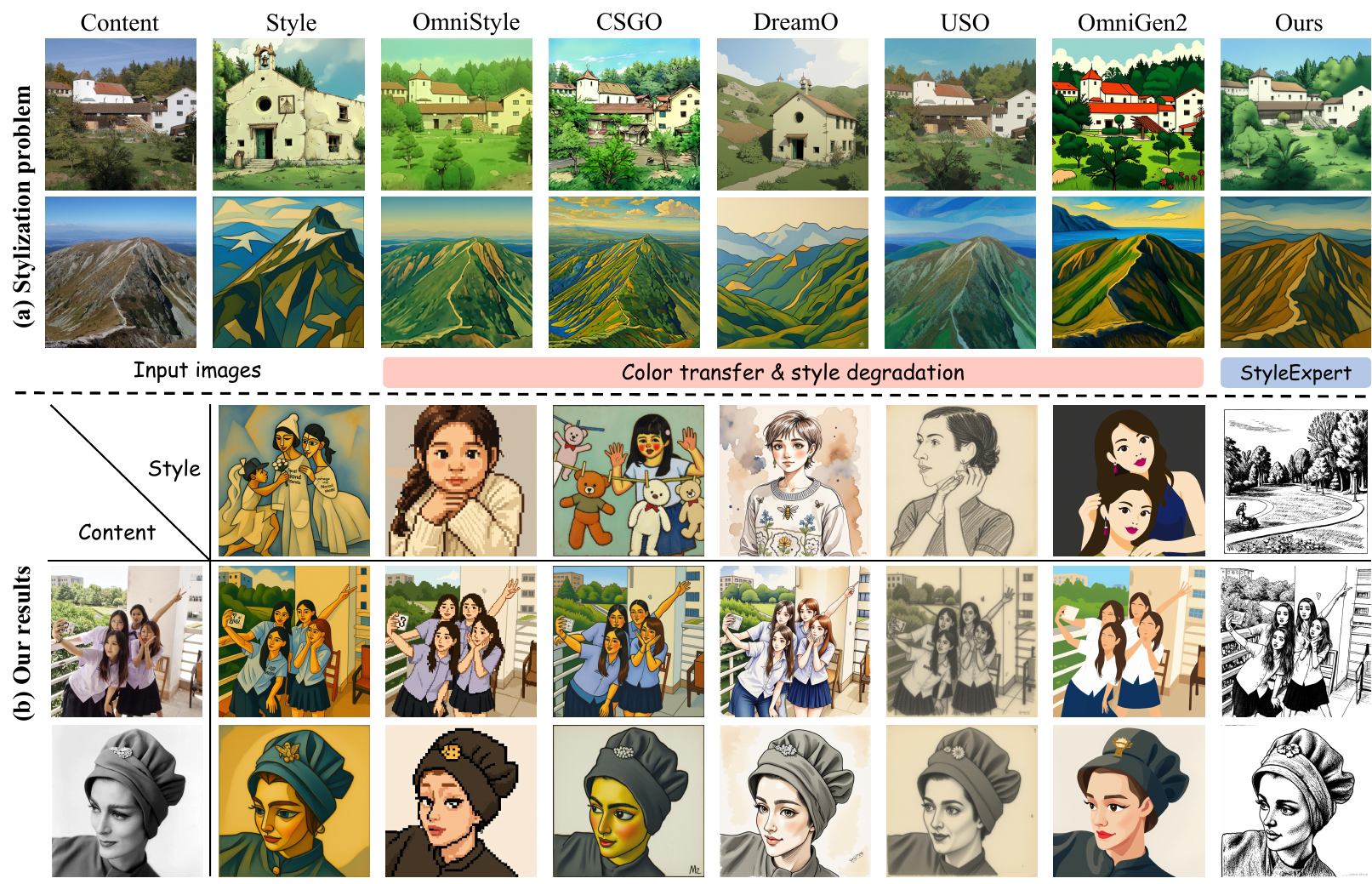}    
  \vspace{-0.6cm}
  \caption{
    (a) We highlight the difference in stylization focus between \ourMthd{} and prior works such as OmniStyle \cite{wang2025omnistyle}, USO \cite{wu2025uso}, and OmniGen2 \cite{wu2025omnigen2}. 
    While these methods often perform simple color transfer, our approach emphasizes a semantic understanding of the style image. 
    (b) We showcase \ourMthd{}'s stylization results across diverse style reference images.
  }\label{fig:teaser} 
  \vspace{10pt}  
}]

\renewcommand{\thefootnote}{\fnsymbol{footnote}}
\footnotetext[1]{Corresponding author.}
\renewcommand{\thefootnote}{\arabic{footnote}} 

\begin{abstract}

Diffusion-based stylization has advanced significantly, yet existing methods are limited to color-driven transformations, neglecting complex semantics and material details.
We introduce \ourMthd{}, a semantic-aware framework based on the Mixture of Experts (MoE).
Our framework employs a unified style encoder, trained on our large-scale dataset of content-style-stylized triplets, to embed diverse styles into a consistent latent space.
This embedding is then used to condition a similarity-aware gating mechanism, which dynamically routes styles to specialized experts within the MoE architecture.
Leveraging this MoE architecture, our method adeptly handles diverse styles spanning multiple semantic levels, from shallow textures to deep semantics.  
Extensive experiments show that \ourMthd{} outperforms existing approaches in preserving semantics and material details, while generalizing to unseen styles.
Our code and collected images are available at the project page: \url{https://hh-lg.github.io/StyleExpert-Page/}.

\end{abstract}


\section{Introduction}
\label{sec:intro}

Style transfer \cite{jing2019neural, wu2022completeness, wu2020efanet, cai2023image, xing2024csgo, wang2025omnistyle,lee2024audio}, which aims to alter an image’s aesthetic attributes while preserving its structural content, has rapidly evolved with the emergence of diffusion transformers (DiT) \cite{peebles2023scalable}. 
These models have significantly enhanced the generative quality of stylization, enabling modern diffusion-based frameworks \cite{wang2025omnistyle, wang2025omnistyle2, wu2025uso, mou2025dreamo} to achieve higher efficiency, flexibility, and fidelity.

Depending on the core objective of stylization, style transfer can be categorized into two sub-tasks \cite{jing2019neural}: \textit{color transfer} and \textit{semantic transfer}.
Color transfer \cite{wang2025omnistyle, xing2024csgo} aims to adapt the color distribution of the style reference to the content image while preserving its spatial structure.
In contrast, semantic transfer \cite{zhou2025attention, qin2025free} emphasizes transferring texture, line, and material from the style image to the content image, sometimes allowing subtle spatial adjustments to better reflect the desired stylistic expression.
As visually demonstrated in \figref{fig:teaser}(a), prominent state-of-the-art methods \cite{wang2025omnistyle, xing2024csgo, wu2025omnigen2, mou2025dreamo, wu2025uso, zhou2025attention} often degenerate into simplistic color mapping. 
We observe that most methods merely transfer dominant colors from the style image, such as green or yellow, and fail to capture the core semantic elements (\eg, texture, brushstrokes).
This deficiency in existing methods is mainly due to two issues: \emph{color and semantic imbalance in existing style transfer datasets} and \emph{limitations in style information integration methods.}

Existing style transfer datasets \cite{wang2025omnistyle, xing2024csgo} exhibit a significant imbalance in style categories, predominantly emphasizing color-based styles while under-representing semantic and material styles.
Although style datasets like Style30K \cite{li2024styletokenizer, schuhmann2022laion} include a fair amount of texture-rich references, the training-free style transfer methods \cite{wang2024instantstyle, Chung_2024_CVPR, frenkel2024implicit, gao2025styleshot, wang2024plus} commonly used for stylization generation often yield low-quality stylized images contaminated with irrelevant textures, noise, and visual artifacts.
This imbalance, combined with the suboptimal performance of earlier stylization methods, leads to datasets with poor image quality.

For the second issue, existing style transfer methods, such as OmniStyle \cite{wang2025omnistyle} and DreamO \cite{mou2025dreamo}, inject style information into the model by concatenating VAE \cite{kingma2013auto} latent codes.
However, due to the limited semantic information in the VAE latent space, it is challenging to capture and apply high-level semantic features inherent in the style image.
Other methods, such as CSGO \cite{xing2024csgo} and USO \cite{wu2025uso}, incorporate style images via cross-attention or prompts, but they fail to account for the diversity of semantic characteristics in style images, treating all styles uniformly during injection into the model.

To address the aforementioned issues, we first construct a semantic-diverse style dataset.
We address the color and semantic imbalance in existing datasets by leveraging style-centric LoRA models from the Hugging Face community, which capture the diverse high-level semantic styles missing from these datasets.
Our pipeline begins by manually selecting these LoRAs and applying them, along with the content-preserving OmniConsistency LoRA \cite{song2025omniconsistency}, to stylize content images.
To form the content-style-stylized data triplet, we use CLIP \cite{radford2021learning} to select the most representative style reference image from the style's domain.
Finally, we employ Qwen \cite{qwen} as a quality filter to remove artifact-contaminated triplets.

To incorporate style information into the model in a semantically rich manner, with distinct approaches for each style, we first attempt to use pre-trained community LoRA in combination with a style selector. 
However, this method suffers from poor scalability for both the style selector and the pre-trained LoRA, as they are trained independently and are difficult to integrate. 
As an alternative, we propose jointly training these style LoRAs, transforming the problem into a classic Mixture of Experts (MoE) structure \cite{dai2024deepseekmoe}. 
To accelerate MoE convergence and enhance the network's generalization to styles, we pre-train a style encoder with the InfoNCE loss \cite{oord2018representation} and integrate it into the router. 
This enables the router to effectively recognize different styles and map images with similar styles to adjacent latents, further improving the network’s generalization ability and the stability of MoE training in its early stage. 
As shown in \figref{fig:teaser}(a), our method demonstrates superior performance by faithfully capturing the color palette, line work, and overall atmosphere.

In summary, our contributions are as follows: 
\begin{itemize}
\item We propose a novel pipeline optimized for semantic style transfer, which uses a single style image to extract semantic information and generate high-fidelity, content-preserving stylization.

\item We introduce a novel method that employs InfoNCE loss to train a style encoder, which is integrated into the MoE router. 
This approach enables fast, stable convergence during MoE training while improving the network's generalization across various styles.

\item We construct a dataset of 500k content-style-stylized triplets for high-quality research on semantic stylization and customization.

\end{itemize}

\begin{figure*}[t]
  \centering   
  \includegraphics[width=0.9\linewidth]{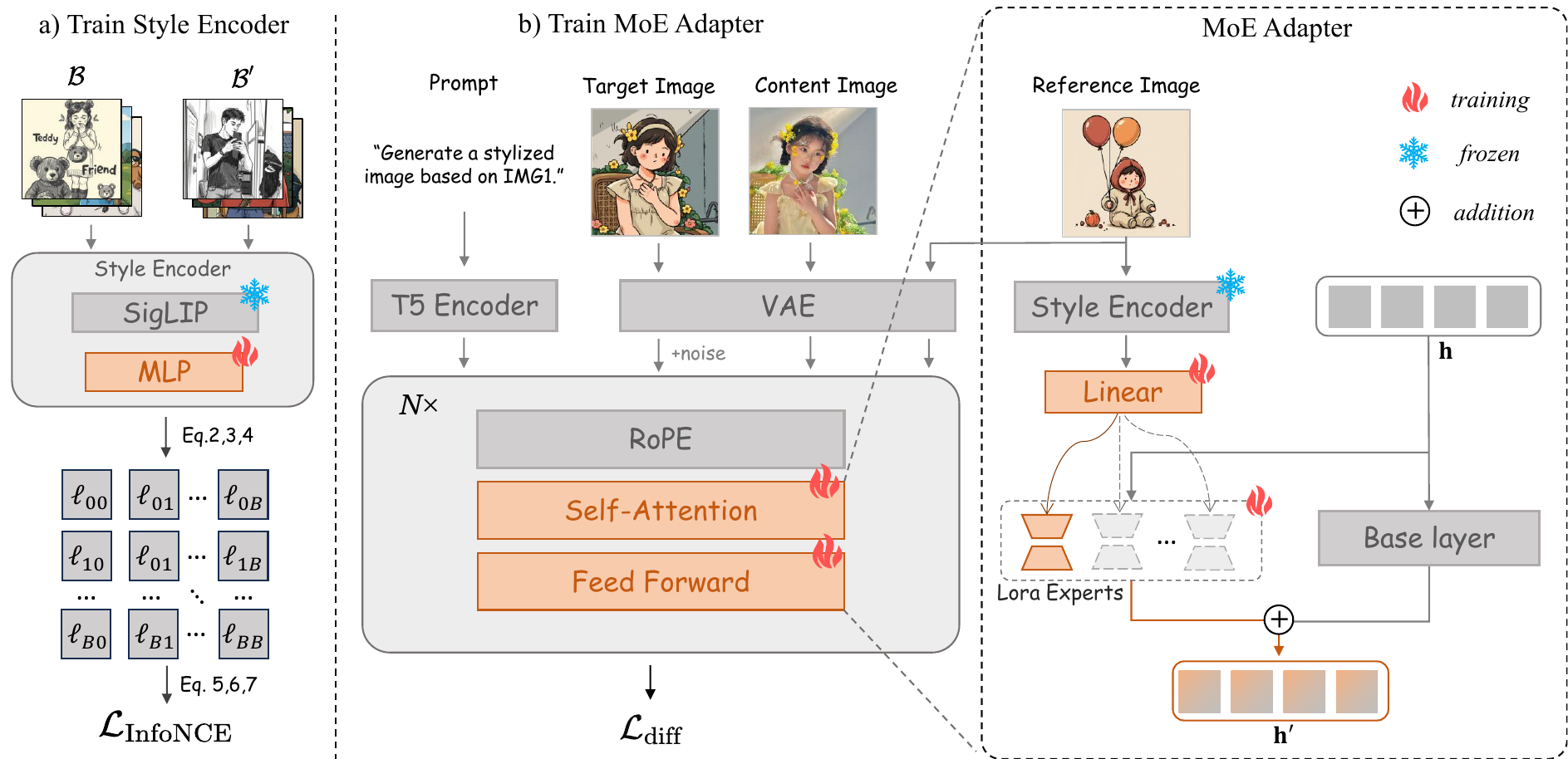}
  \caption{Overview of the proposed \ourMthd{}. 
    Our method comprises two training stages. 
    In the first stage (\secref{sec:style_encoder}), a style encoder is trained with the InfoNCE loss to learn discriminative style representations, thereby accelerating convergence. 
    In the second stage (\secref{sec:moe}), the pre-trained encoder provides style priors to guide the router network in training MoE LoRA adapters, enabling each layer to dynamically select the most suitable experts for diverse styles.
  }\label{fig:method}
  \vspace{-10pt}
\end{figure*}

\section{Related Work}
\label{sec:relatedwork}

\myPara{Style Transfer.}
Style transfer aims to transfer the style from a reference image to a target image.
Early works on stylization primarily focused on optimization-based methods \cite{gatys2016image, gatys2017controlling, kolkin2019style}, leveraging feature properties to achieve stylization. 
Following the recent success of diffusion models such as Stable Diffusion \cite{esser2024scaling,podell2023sdxl} and FLUX \cite{flux2024} in the text-to-image domain, a growing number of diffusion-based stylization methods have been proposed.

Among recent diffusion-based style transfer methods, numerous training-free approaches \cite{chung2024style, jeong2024training, gao2025styleshot, sohn2023styledrop, wang2024instantstyle, wang2024plus, xing2024csgo, xu2024freetuner, zhang2023inversion,liu2025sad} have emerged. 
Prominent examples include B-LoRA \cite{frenkel2024implicit}, K-LoRA \cite{ouyang2025k}, and Attention Distillation \cite{zhou2025attention}, which inject style information either by leveraging pre-trained style LoRAs or by performing optimization during inference.
These training-free methods often suffer from unstable performance and practical limitations, such as inference-time computational overhead or the inability to use a single style image.
Consequently, training-based methods have gained popularity.
Most of these methods, including OmniStyle \cite{wang2025omnistyle}, DreamO \cite{mou2025dreamo}, and CSGO \cite{xing2024csgo}, train an adapter or an extra conditional branch to inject style and content information, adopting strategies similar to ControlNet \cite{zhang2023adding} and EasyControl \cite{zhang2025easycontrol}.
Other methods, such as USO \cite{wu2025uso}, integrate style information in a multi-modal fashion by injecting image tokens into the prompt.

Despite recent progress, image stylization methods are challenged by the inherent complexity of artistic styles.
Styles encompass diverse attributes, from pixel-level color to semantic-level properties like texture, lines, and ambiance. 
Consequently, existing approaches often fail to satisfy the specific requirements of all style types.

\myPara{Mixture of Experts.}
Mixture of Experts (MoE) models \cite{Jacobs1991moe, jordan1994hierarchical, shazeer2017outrageously} are renowned for their ability to increase model capacity through parameter expansion.
Technically, a MoE layer \cite{fedus2022switch, shazeer2017outrageously} consists of $N$ expert networks and a router network to activate a subset of these experts and combine their outputs.
A notable direction focuses on combining MoE with LoRA \cite{zhang2025context, yang2024multi}, employing a sparse top-k expert routing mechanism to maintain efficiency while augmenting capacity across various tasks.

However, the exploration of MoE architectures in image generation remains limited.
Existing MoE-finetuned models, such as ICEdit \cite{zhangenabling} and MultiCrafter \cite{wu2025multicrafter}, utilize LoRA as experts and feed the hidden states as conditional input to the router. 
In contrast to these approaches, we pre-train a style encoder to provide style priors. 
We then feed the encoder's latent features extracted from a style image to the router as a conditional input to control expert selection.

\section{Method}
    
In this section, we first introduce the fundamentals of the DiT model, which serve as the basis for our proposed method.
We then present our method framework in detail, consisting of two training stages, as illustrated in \figref{fig:method}.
In the first stage, we train a style representation encoder using the InfoNCE loss \cite{oord2018representation} to ensure its generalization ability across various styles, providing a foundation for subsequent training.
In the second stage, we use the prior knowledge from the trained style representation encoder to inform the MoE router, enabling rapid convergence during MoE training.

\subsection{Preliminaries}
\label{sec:Preliminaries}

Following DiT \cite{peebles2023scalable}, we use a multimodal attention mechanism that combines text and image embeddings. 
This operation is defined as:
\begin{equation} 
  \text{Attention}(Q, K, V) = \text{softmax}\left( \frac{Q K^\top}{\sqrt{d}} \right) V, 
\end{equation}
where the query $Q$, key $K$, and value $V$ representations are linear projections of the input tokens $Z$ (\ie, $Q = W_Q Z$, $K = W_K Z$, $V = W_V Z$).
For this architecture, the input $Z = [c, z_t]$ represents the concatenation of the text token $c$ and the noisy image token $z_t$. 
Recent models, such as the Flux-Kontext \cite{labs2025flux1kontextflowmatching, flux2024} image editing model, support $Z' = [c, z_t, z_c]$, which facilitates the inclusion of image control $z_c$. 
Given this architectural flexibility to support image control inputs, along with its proven strength in image editing, we adopt the Flux-Kontext as our base model.

\begin{figure*}[t]
  \centering   
  \includegraphics[width=\linewidth]{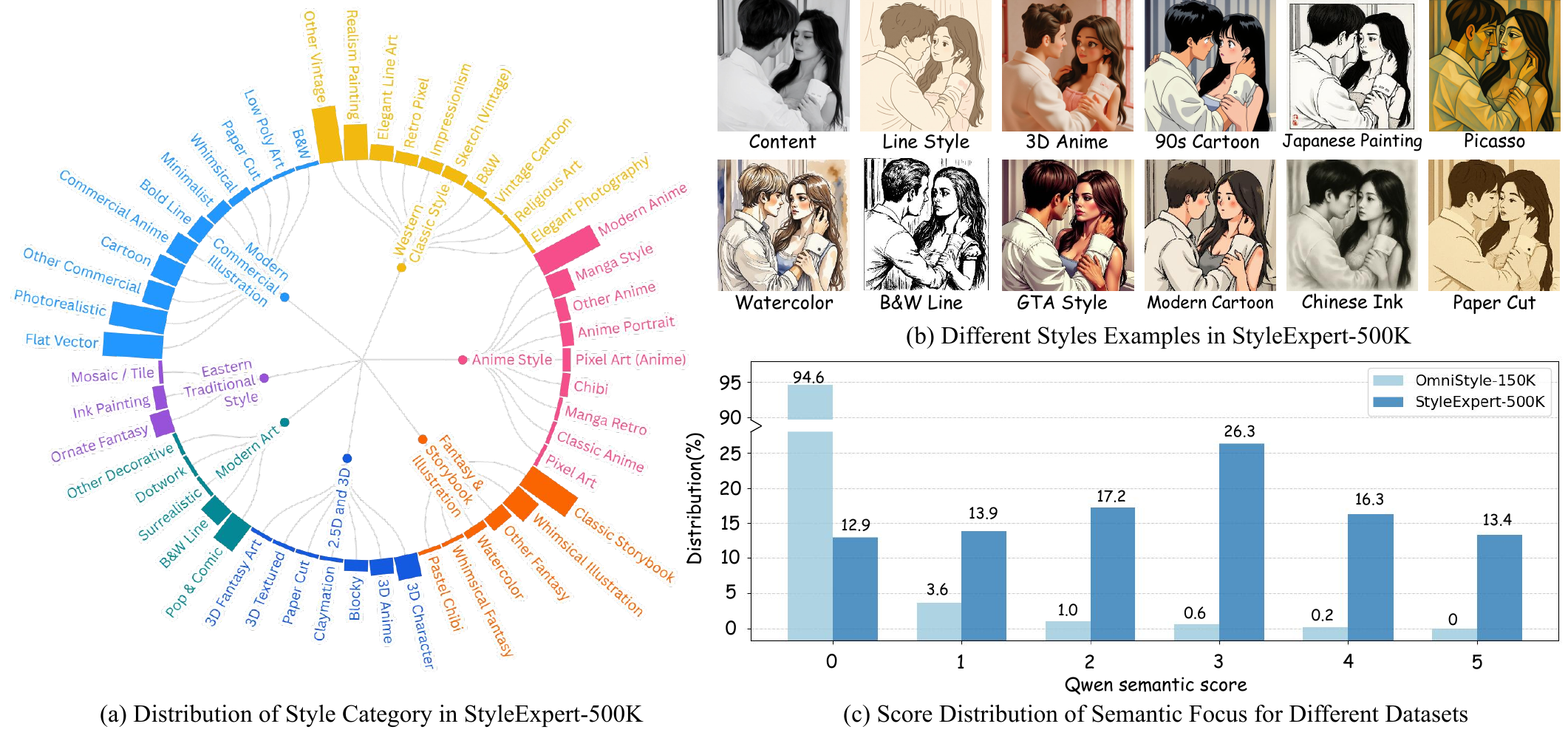}
  \caption{Overview of \ourMthd{}-500K. 
    (a) illustrates the hierarchical distribution of all 209 styles in \ourMthd{}-500K, where bar heights indicate the number of styles per category. 
    (b) presents examples of nine different stylizations for the same content from our \ourMthd{}-500K dataset. 
    (c) compares the focus on semantic stylization between \ourMthd{}-500K and OmniStyle-150K.
  }\label{fig:dataset_overview}
  \vspace{-8pt}
\end{figure*}

\subsection{Style Representation Encoder}
\label{sec:style_encoder}

In this subsection, we fine-tune a model to extract style representations.
Given an image $I_i$ with style label $s_i$, our objective is to learn a representation $\mathbf{e}_i$ such that the distance $d(\mathbf{e}_i, \mathbf{e}_j)$ is minimized for any pair of images $(I_i, I_j)$ sharing the same style label ($s_i = s_j$).
We employ a temperature-scaled cosine similarity as our metric $d(\cdot, \cdot)$:
\begin{equation}\label{eq:logits}
d(\mathbf{e}_i, \mathbf{e}_j) = \frac{\mathbf{e}_i \cdot \mathbf{e}_j^T}{\tau \|\mathbf{e}_i\| \|\mathbf{e}_j\|},
\end{equation}
where $\tau$ is a temperature parameter.
We compute the style representation $\mathbf{e}_i$ by passing features from a pre-trained SigLIP \cite{zhai2023sigmoid} model through an MLP network $\Phi$.
Following \cite{zhang2024ssr}, we concatenate the hidden states $\{\mathbf{h}_i^{(l)}\}_{l=1}^L$ from $L$ different SigLIP layers:
\begin{equation}\label{eq:sty_latent}
    \mathbf{e}_i = \Phi(\text{concat}(\mathbf{h}_i^{(1)}, \mathbf{h}_i^{(2)}, \dots, \mathbf{h}_i^{(L)})).
\end{equation}
We train the MLP $\Phi$ using an InfoNCE contrastive loss \cite{oord2018representation}, similar to CLIP \cite{radford2021learning}.
To form positive and negative pairs, we compute the loss between two independently sampled batches, $\mathcal{B}$ and $\mathcal{B}'$.
Let $\mathbf{E} = \{\mathbf{e}_i\}_{i=1}^B$ and $\mathbf{E}' = \{\mathbf{e}'_j\}_{j=1}^B$ be their corresponding style representations.
We compute a $B \times B$ matrix of log-probabilities $\ell$, where each element $\ell_{ij}$ compares $\mathbf{e}_i \in \mathbf{E}$ against all representations in $\mathbf{E}'$ via a softmax:
\begin{equation}
    \ell_{ij} = \log \left( \frac{\exp(d(\mathbf{e}_i, \mathbf{e}'_j))}{\sum_{k=1}^B \exp(d(\mathbf{e}_i, \mathbf{e}'_k))} \right).
\end{equation}

We then define a positive mask $\mathbf{M} \in \mathbb{R}^{B \times B}$, which compares the style label $s_i$ from the first batch with the style label $s'_j$ from the second batch. The mask $\mathbf{M}_{ij}$ is 1 if the images share the same style label, and 0 otherwise:
\begin{equation}
\mathbf{M}_{ij} = \begin{cases}
1, & \text{if } s_i = s'_j, \\
0. & \text{if } s_i \neq s'_j.
\end{cases}
\end{equation}
This mask is crucial for weighting log probabilities based on whether images from the two batches belong to the same style.

The InfoNCE loss for each sample in the first batch is computed by summing over all pairs of images, weighted by the positive mask $\mathbf{M}$. The loss per sample for image $I_i$ can be formulated as:
\begin{equation}
    \mathcal{L}_i = -\frac{1}{\sum_{j=1}^B \mathbf{M}_{ij}} \sum_{j=1}^B \mathbf{M}_{ij} \cdot \ell_{ij}.
\end{equation}
Finally, we compute the overall loss by averaging the individual losses across the entire batch:
\begin{equation}
    \mathcal{L}_{\text{InfoNCE}} = \frac{1}{B} \sum_{i=1}^B \mathcal{L}_i.
\end{equation}
This final loss term is used to train the MLP $\Phi$, encouraging the model to learn meaningful style representations that are consistent within each style label.

\subsection{Efficient MoE Fine-tuning for Style Transfer}
\label{sec:moe}

While LoRA fine-tuning \cite{wang2025omnistyle,wang2025omnistyle2} is used for stylization, a single LoRA model cannot handle diverse styles at varying granularities. 
To overcome this, we adopt a Mixture of Experts (MoE) framework that uses a router to select the most suitable experts for each style.

In particular, we incorporate style references into the DiT network by embedding $N_e$ LoRA experts within both the self-attention layers and the FFN linear layers. 
Let $\mathbf{h} \in \mathbb{R}^{d_{\text{in}}}$ denote the input to a given layer $l$. 
The output of this layer is computed as follows.
For the style reference image $I_s$, we use its style latent $\mathbf{e_s}$, calculated using \eqref{eq:sty_latent}, as the condition latent input to the router. 
This router then assigns weights to each expert. 
The weight for the $i$-th expert is given by:
\begin{equation}
w_i = \text{softmax}(\text{TopK}(g(\mathbf{e}_{s}), k))_i,
\end{equation}
where $g(\mathbf{e}_{s})$ is the output of the router function. 
The $\text{TopK}$ operation selects the top $k$ values from $g(\mathbf{e}_{s})$ and assigns $-\infty$ to others. 

Finally, the output $\mathbf{h}'$ of this layer is obtained by combining the original transformation with contributions from both a shared expert and the selected specialized experts:
\begin{equation}
\mathbf{h}' = l(\mathbf{h}) + \frac{\alpha}{r} (B_s \cdot A_s  + \sum_{i=1}^{N_e} w_i \cdot B_i \cdot A_i)\cdot \mathbf{h},
\end{equation}
where $l(\mathbf{h}) \in \mathbb{R}^{d_{\text{out}}}$ is the original output. 
$B_s \in \mathbb{R}^{d_{\text{out}} \times r}$ and $A_s \in \mathbb{R}^{r \times d_{\text{in}}}$ are the LoRA \cite{hu2022lora} weights for the shared expert, while $B_i$ and $A_i$ are the weights for the $i$-th specialized expert. 
$\alpha$ is the scaling factor and $r$ represents the LoRA rank. 
The combination of the shared expert and the weighted sum of specialized experts enhances the model’s capacity to effectively capture a diverse range of styles.

\section{Stylized Dataset Curation}
\label{dataset}

To construct a dataset with better-balanced color and semantics, we first evaluate the existing OmniStyle-150K \cite{wang2025omnistyle}.
We assess the feasibility of its data-generation pipeline, which leverages current SOTA style transfer methods.
To this end, we benchmark each style in the dataset using a Qwen Semantic Score (detailed in the Supplementary Material), specifically focusing on whether the transfer prioritizes semantic information, such as texture and material, over superficial color features.
Our evaluation results, as shown in \figref{fig:dataset_overview}(c), reveal that the vast majority of styles in OmniStyle-150K—841 out of 889—overwhelmingly focus on simple color transfer. 
This finding highlights the need for a new paradigm to create more balanced, semantically rich style-transfer datasets that are independent of existing SOTA stylization techniques.

\begin{figure*}[t]
  \centering   
  \includegraphics[width=\linewidth]{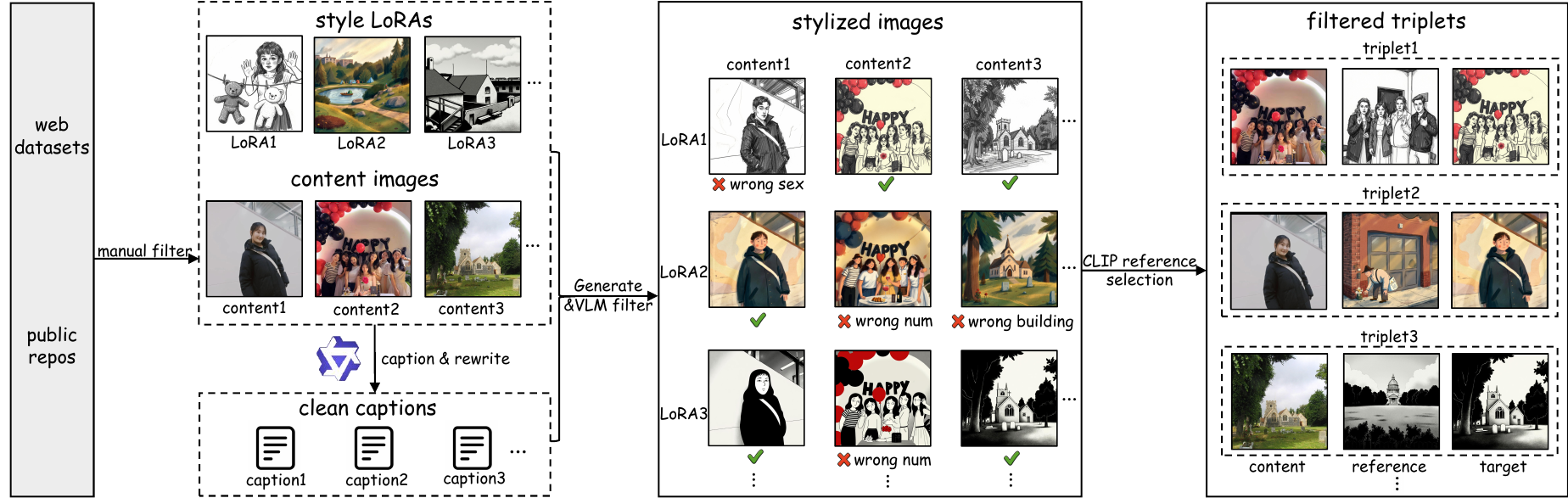}
  \caption{
    Overview of our dataset curation pipeline. 
    We first manually collect content images and style LoRAs from the web. 
    For each content image, we use Qwen \cite{qwen} to generate a clean caption that excludes color and style information. 
    The content image, its caption, and the corresponding style LoRA are then fed into the model to generate stylized images. 
    A vision-language model (VLM) filters out those with incorrect or failed stylization. 
    Finally, we compute CLIP \cite{radford2021learning} similarity to select the most suitable style reference for each target, forming the final triplet dataset.
  }\label{fig:dataset_pipeline}
  \vspace{-10pt}
\end{figure*}

\figref{fig:dataset_pipeline} illustrates our data generation pipeline. 
To address the scarcity of semantic styles in existing datasets, we leverage style LoRAs from the Hugging Face community, as they provide a rich source of both pixel-level and semantic-level styles. 
We initially collected approximately 650 style LoRAs. 
To mitigate inconsistent quality, we implemented a rigorous filtering process of manual curation and de-duplication, resulting in a refined set of 209 high-quality LoRAs.

To generate highly diverse content, we curated a base image collection of approximately 2,700 photographs.
This collection spans a wide range of categories, including people, landscapes, architecture, animals, and even complex multi-person scenes, ensuring varied content for stylization. 
We generated descriptive captions for all images. 
However, to prevent these captions from introducing confounding stylistic information (\eg, atmosphere, lighting) unrelated to the target LoRA, we utilized Qwen \cite{qwen} to rewrite them. 
This refinement ensures the prompts describe only the objective content, allowing the style LoRA to apply a consistent stylization without interference.

To stylize the content images, we employ the OmniConsistency LoRA \cite{song2025omniconsistency}.
We combine the content image, its corresponding clean prompt, and a style LoRA to generate compositionally consistent stylized results.
This process yielded approximately 500,000 images, which comprise our new dataset: \ourMthd{}-500K.
\figref{fig:dataset_overview}(a) and \figref{fig:dataset_overview}(b) provide qualitative examples illustrating the diverse style types within our dataset.
As shown in \figref{fig:dataset_overview}(c), \ourMthd{}-500K achieves a better balance between color-centric and semantic-centric styles compared to OmniStyle-150K.

To enhance the semantic and spatial consistency between the synthesized images and their original content counterparts, we introduce an additional filtering step.
We employ Qwen-VL \cite{qwenvl} to prune results exhibiting poor stylization, significant layout degradation, incorrect demographic attributes (\eg, age, gender), or object inconsistencies.
This rigorous curation yields our final dataset of around 40,000 high-fidelity images, which we name \ourMthd{}-40K.

Finally, to construct the triplets of \emph{(content image $I_c$, style image $I_s$, stylized image $I_{sc}$)}, we select an appropriate style reference $I_s$ for each generated image $I_{sc}$.
We designate the set of all stylized images for a single style as $\mathcal{D}_{sc} = \{I_{sc}^{(i)}\}_{i=1}^N$.
For each individual image $I_{sc}^{(i)} \in \mathcal{D}_{sc}$, we select its style reference $I_s$ by finding the most visually similar image $I_{sc}^{(k)}$ from this same set, under the constraint that $I_{sc}^{(k)} \neq I_{sc}^{(i)}$.
This selection is performed by computing the CLIP-based similarity:
\begin{equation}
I_s^{*} = \arg\max_{I_{sc}^{(k)} \neq I_{sc}^{(i)}} \mathrm{CLIPSim}\big(I_{sc}^{(i)}, I_{sc}^{(k)}\big),
\end{equation}
where $\mathrm{CLIPSim}(\cdot,\cdot)$ denotes the cosine similarity between CLIP embeddings. This process yields a coherent triplet $(I_c, I_s^{*}, I_{sc}^{(i)})$ where the style reference is itself a generated example of the style.

\begin{figure*}[t]
    \centering   
    \setlength{\abovecaptionskip}{2pt}
    \includegraphics[width=\linewidth]{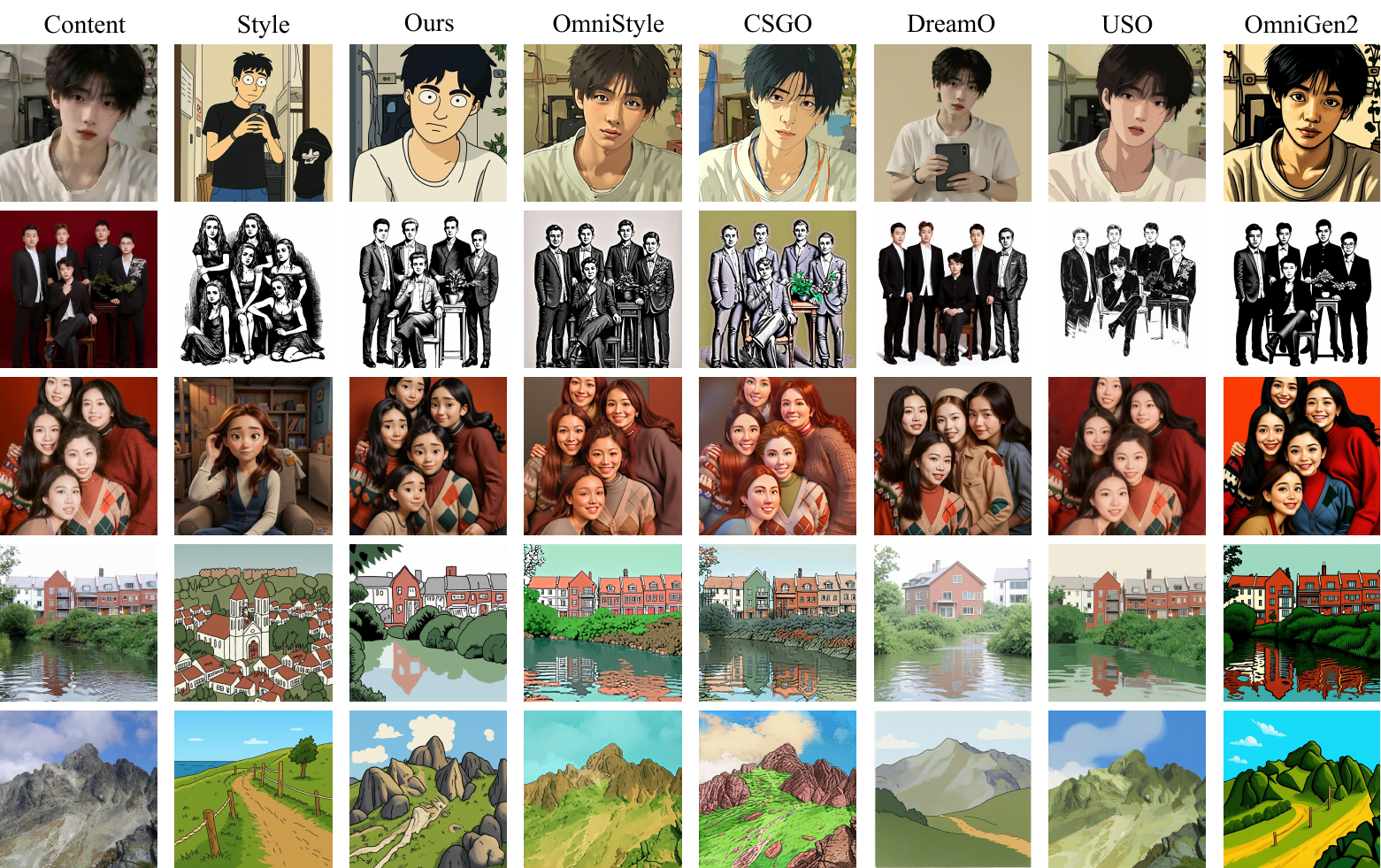}
    \caption{Qualitative comparison of our method, \ourMthd{}, with other SOTA style transfer methods on unseen styles.}
    \label{fig:quality_unseesn}
    \vspace{-8pt}
\end{figure*}

\section{Experiments}

\subsection{Experiment Settings}

\myPara{Baselines.} 
We selected recent stylization methods for comparison, all of which support multi-image inputs (a content image and a style image).
These methods are: OmniStyle \cite{wang2025omnistyle}, CSGO \cite{xing2024csgo}, DreamO \cite{mou2025dreamo}, Qwen-Image-Edit \cite{wu2025qwenimagetechnicalreport}, and OmniGen2 \cite{wu2025omnigen2}.

\myPara{Benchmark.} 
To fairly compare our method with others, we use 90\% of the styles for training and 10\% for testing, with 188 styles in the training set and 21 in the test set.
The style encoder and MoE fine-tuning are trained only on the training set.
For the test set, we randomly select 50 pairs of content and style images per style, generating two images per pair with different seeds, yielding 2,100 images per method.

\myPara{Evaluation Metrics.} 
We evaluate all methods across three dimensions: content fidelity, stylization degree, and aesthetic quality.
For content fidelity, we use CLIP \cite{radford2021learning} and DINO \cite{zhang2022dino} scores.
For style similarity, we employ CSD \cite{somepalli2024measuring} and DreamSim \cite{fu2023dreamsim}.
For aesthetic quality, we adopt the LAION aesthetic score \cite{laionaes}.
Furthermore, to highlight the advantages of semantic-level stylization, we employ the Qwen Semantic Score (detailed in the Supplementary Material) to measure attributes such as material and line style.

\myPara{Implementation details.} 
To train the style representation encoder, we employ the AdaBelief optimizer with a learning rate of $1\mathrm{e}{-5}$ and a batch size of $128$ for $3500$ steps.
For MoE LoRA adapter fine-tuning, we use Flux Kontext \cite{flux2024,labs2025flux1kontextflowmatching} as our base model.
We set the LoRA rank to $8$ for each of the $16$ experts. 
We select the top $2$ experts per layer using a batch size of $1$ per GPU (total $4$ with $4$ GPUs) and a learning rate of $1\mathrm{e}{-4}$ for $10,000$ iterations.

\begin{table}[t]
\centering
\scriptsize
\setlength{\tabcolsep}{2pt}
\caption{Quantitative comparison of our method against other stylization methods, as well as LoRA-only and MoE-only fine-tuning baselines. Best results are marked in \textbf{bold} and second-best results are \underline{underlined}. All metrics except for the Aesthetic score are presented as percentages(\%).}
\begin{tabular}{@{}lcccccc@{}}
\toprule
                   & CLIP & DINO & CSD  & Aesthetic & Qwen Semantic & DreamSim \\ 
Method                     & ($\uparrow$) & ($\uparrow$) & ($\uparrow$)  & ($\uparrow$) & ($\uparrow$) & ($\downarrow$) \\ \midrule
CSGO                       & 63.41          & 65.50          & 61.07          & 6.28                & 28.93                   & 42.39                \\
DreamO                     & 64.14          & 62.68          & 47.91          & 6.20                & 19.29                   & 44.95                \\
OmniGen2                   & 67.07          & 63.06          & 55.61          & 6.13                & 23.69                   & 41.55                \\
OmniStyle                  & 65.39          & {\ul 72.27}    & 59.65          & 6.07                & 40.00                   & 41.83                \\
Qwen-Image-Edit            & 67.47          & 55.86          & 56.74          & 6.20                & 42.74                   & {\ul 34.47}          \\
USO                        & {\ul 69.39}    & \textbf{84.03} & 53.60          & 6.30                & 19.88                   & 48.62                \\ \midrule
LoRA Training              & 67.33          & 60.78          & {\ul 70.88}    & 6.13                & {\ul 70.71}             & 36.77                \\
MoE Training               & 67.83          & 62.48          & 66.70          & {\ul 6.43}       & 71.43                   & 38.54                \\ \midrule
\textbf{\ourMthd{}(Ours)} & \textbf{70.19} & 64.72          & \textbf{73.18} & \textbf{6.48}          & \textbf{75.12}          & \textbf{28.18}       \\ \bottomrule
\end{tabular}
\vspace{-10pt}
\label{tab:quantity}
\end{table}

\subsection{Qualitative Comparisons}

\figref{fig:quality_unseesn} presents a qualitative comparison of \ourMthd{} against competing approaches on unseen styles.
We observe that our results more faithfully capture the target style reference, particularly in terms of lines (Rows 1, 2, 4), overall atmosphere (Row 3), and materials (Rows 3, 5).
In contrast, existing methods (\eg, OmniStyle, CSGO, USO, and OmniGen2) often degenerate into simple color transfer, failing to capture deeper textural attributes such as line patterns.
Furthermore, some methods, such as DreamO and USO, tend to over-preserve the content image, resulting in poor stylization(Row 3).

\subsection{Quantitative Comparisons}

As shown in \tabref{tab:quantity}, which presents the quantitative comparison, our method achieves state-of-the-art results on the CLIP Score, CSD Score, Aesthetic Score, Qwen Semantic Score, and DreamSim metrics.
This demonstrates the superiority of our approach in maintaining both style consistency and content fidelity.
Regarding the lower DINO score, we posit that it penalizes our method's successful transfer of material-altering semantic styles.
Competing methods often fail at this, defaulting to mere color transfer, which preserves the original material and thus achieves a deceptively higher DINO score.
Notably, our method's Qwen Semantic Score (75.12) significantly surpasses all competing methods by a large margin.
This result strongly indicates that our approach excels at transferring complex semantic style information, thereby validating the effectiveness of both our data pipeline and our method.

\begin{table}[t]
\centering
\setlength{\abovecaptionskip}{2pt}
\small
\caption{Computation and Parameter Comparison} 
\begin{tabular}{@{}lcc@{}}
\toprule
    & Computation (G) & Trainable Params (M) \\ \midrule
Base Model          & 10.92                & -                    \\
+ LoRA       & +0.67                & 751.48              \\
+ MoE Experts  & +0.12                & 818.71              \\ \bottomrule
\end{tabular}
\vspace{-5pt}
\label{tab:computation_comparison} 
\end{table}

\subsection{Ablation Study}

\begin{figure}[tbp]
    \centering
    \setlength{\abovecaptionskip}{2pt}
    \includegraphics[width=1\linewidth]{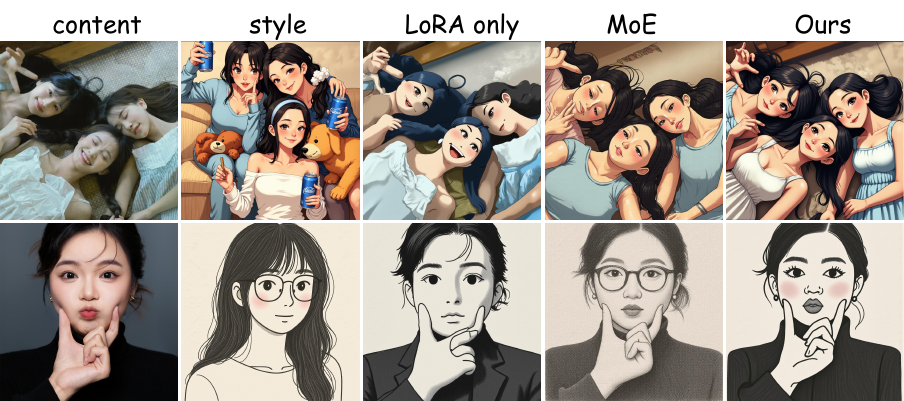}
    \caption{Qualitative comparison of our method against LoRA-only and MoE-only fine-tuning baselines.}
    \label{fig:ablation_quality}
    \vspace{-6pt}
\end{figure}

\myPara{Qualitative comparison.} \figref{fig:ablation_quality} qualitatively compares our method against the LoRA-only and MoE-only (without pre-trained style encoder) baselines.
As shown, the LoRA-only baseline struggles with complex styles, failing to capture semantic information.
The MoE-only baseline is unstable, resulting in under-stylization or erroneous content transfer (\eg, adding glasses from the style image in the second row).
In contrast, our full method consistently achieves the highest-quality stylization.

\myPara{Quantitative comparison.} 
The last three rows of \tabref{tab:quantity} quantitatively ablate our method against LoRA-only and MoE-only fine-tuning.
Without the pre-trained style encoder, the MoE baseline's performance degrades, particularly on the CSD and DreamSim metrics, where it underperforms standard LoRA fine-tuning.
We attribute this degradation to the inherent instability of MoE training.
In contrast, our full method achieves the best performance across all key metrics for content fidelity and style similarity.

\myPara{Efficiency.} 
\tabref{tab:computation_comparison} presents a comparison of computational load and trainable parameters per step between our method and standard LoRA, given a consistent total LoRA rank.
Our method incurs less computational overhead on the base model, leading to superior computational efficiency.
Conversely, the increased number of trainable parameters suggests that our MoE approach has greater capacity for knowledge storage than standard LoRA.

\begin{figure}[tbp]
    \centering
    \setlength{\abovecaptionskip}{2pt}
    \includegraphics[width=1\linewidth]{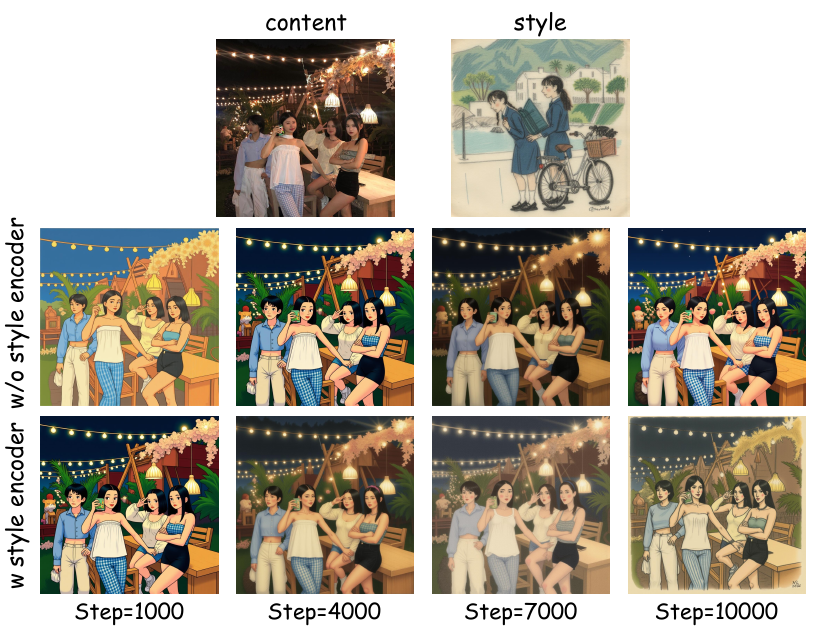}
    \caption{Impact of the style encoder on MoE Training. }
    \label{fig:ablation_sty_encoder_instable}
    \vspace{-5pt}
\end{figure}

\myPara{Convergence Speed.} 
As shown in \figref{fig:ablation_sty_encoder_instable}, removing the style encoder leads to optimization instability and slower convergence for the MoE architecture.
We hypothesize this is because the MoE training is inherently unstable at the beginning, lacking a strong signal to guide the router.
This can lead to the selection of incorrect or suboptimal experts.
The inclusion of our pre-trained style encoder provides crucial guidance, thereby stabilizing the training process and accelerating model convergence.

\begin{figure}[t]
    \centering
    \setlength{\abovecaptionskip}{2pt}
    \includegraphics[width=0.9\linewidth]{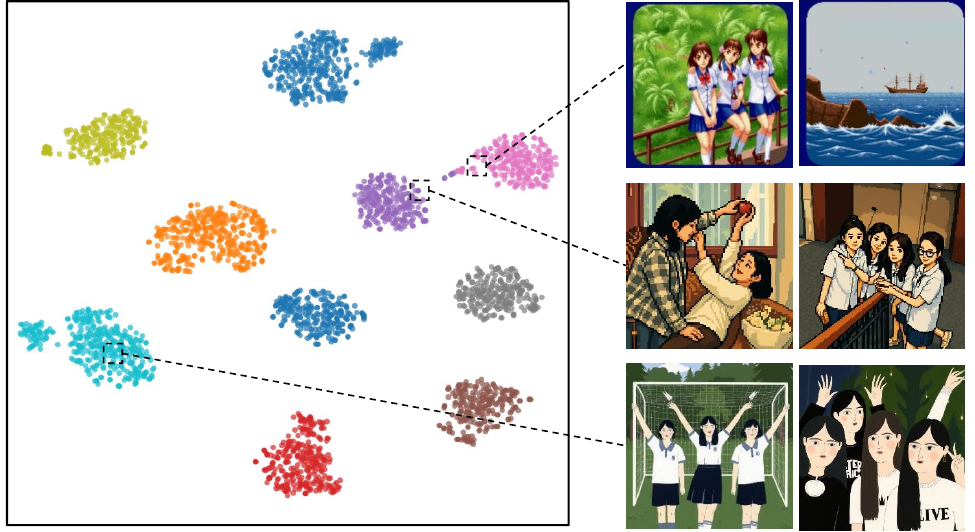}
    \caption{t-SNE visualization of style embeddings from our style encoder on the test set.}
    \label{fig:sty_cluster}
\end{figure}

\begin{table}[tbp]
\centering
\setlength{\abovecaptionskip}{2pt}
\setlength{\tabcolsep}{18pt}
\small
\caption{
    Analysis of MoE expert routing overlap (IoU) between similar styles and dissimilar styles generation. 
}
\label{tab:moe_overlap_compact}
\begin{tabular}{@{}lcc@{}}
\toprule
Model Stage & Similar Styles & Dissimilar Styles \\
\midrule
Early Stage & 34.01\% & 9.77\% \\
Mid Stage   & 34.36\% & 10.32\% \\
Late Stage  & 31.15\% & 11.70\% \\
\midrule[0.5pt]
\textbf{Overall Avg.} & \textbf{33.18\%} & \textbf{10.60\%} \\
\bottomrule
\end{tabular}
\label{tab:expert_overlap}
\end{table}

\section{Discussions and Conclusions}

To investigate how the style encoder functions within the MoE framework, we visualize the latent representations extracted by the style encoder from test images of different styles, as shown in \figref{fig:sty_cluster}.
The results demonstrate that our style encoder generalizes well to unseen styles.
It maps visually similar styles—such as Contra and classic pixel art—closer in the latent space, while mapping illustration-style images farther apart.
This indicates that the style encoder effectively encodes similar styles into nearby latent representations, facilitating generalization to novel styles.

We then explicitly validate this router guidance by analyzing expert selection.
We fix the content of the reference images and use the latent codes produced by the style encoder as conditions for the router network to select experts. 
As shown in \tabref{tab:expert_overlap}, the overlap of selected experts among similar styles is three times higher than that of dissimilar styles.
This confirms that the style encoder's structured latent space successfully guides the router network to assign similar styles to similar experts, thereby enabling effective style-specific control during stylization.

We proposed StyleExpert, a novel stylization method based on MoE fine-tuning using a pre-trained style encoder.
Our framework renders diverse styles across multiple semantic levels and generalizes well to unseen styles.
To support this, we construct a new dataset of content-style-stylized triplets, with a more balanced representation of color-centric and semantic-centric styles suitable for a broader range of stylization tasks.
Experiments show that StyleExpert achieves great style fidelity while keeping high content preservation, outperforming recent state-of-the-art approaches.

\myPara{Acknowledge.}
This work is supported by the NSFC (62225604),  Shenzhen Science and Technology Program (JCYJ20240813114237048), “Science and Technology Yongjiang 2035” key technology breakthrough plan project (2025Z053), Chinese government-guided local science and technology development fund projects (scientific and technological achievement transfer and transformation projects) (254Z0102G).

{
    \small
    \bibliographystyle{ieeenat_fullname}
    \bibliography{main}

\begin{thebibliography}{63}
\providecommand{\natexlab}[1]{#1}
\providecommand{\url}[1]{\texttt{#1}}
\expandafter\ifx\csname urlstyle\endcsname\relax
  \providecommand{\doi}[1]{doi: #1}\else
  \providecommand{\doi}{doi: \begingroup \urlstyle{rm}\Url}\fi

\bibitem[Achiam et~al.(2023)Achiam, Adler, Agarwal, Ahmad, Akkaya, Aleman, Almeida, Altenschmidt, Altman, Anadkat, et~al.]{gpt}
Josh Achiam, Steven Adler, Sandhini Agarwal, Lama Ahmad, Ilge Akkaya, Florencia~Leoni Aleman, Diogo Almeida, Janko Altenschmidt, Sam Altman, Shyamal Anadkat, et~al.
\newblock Gpt-4 technical report.
\newblock \emph{arXiv preprint arXiv:2303.08774}, 2023.

\bibitem[Bai et~al.(2023)Bai, Bai, Chu, Cui, Dang, Deng, Fan, Ge, Han, Huang, Hui, Ji, Li, Lin, Lin, Liu, Liu, Lu, Lu, Ma, Men, Ren, Ren, Tan, Tan, Tu, Wang, Wang, Wang, Wu, Xu, Xu, Yang, Yang, Yang, Yang, Yao, Yu, Yuan, Yuan, Zhang, Zhang, Zhang, Zhang, Zhou, Zhou, Zhou, and Zhu]{qwen}
Jinze Bai, Shuai Bai, Yunfei Chu, Zeyu Cui, Kai Dang, Xiaodong Deng, Yang Fan, Wenbin Ge, Yu Han, Fei Huang, Binyuan Hui, Luo Ji, Mei Li, Junyang Lin, Runji Lin, Dayiheng Liu, Gao Liu, Chengqiang Lu, Keming Lu, Jianxin Ma, Rui Men, Xingzhang Ren, Xuancheng Ren, Chuanqi Tan, Sinan Tan, Jianhong Tu, Peng Wang, Shijie Wang, Wei Wang, Shengguang Wu, Benfeng Xu, Jin Xu, An Yang, Hao Yang, Jian Yang, Shusheng Yang, Yang Yao, Bowen Yu, Hongyi Yuan, Zheng Yuan, Jianwei Zhang, Xingxuan Zhang, Yichang Zhang, Zhenru Zhang, Chang Zhou, Jingren Zhou, Xiaohuan Zhou, and Tianhang Zhu.
\newblock Qwen technical report.
\newblock \emph{arXiv preprint arXiv:2309.16609}, 2023.

\bibitem[Bai et~al.(2025{\natexlab{a}})Bai, Cai, Chen, Chen, Chen, Cheng, Deng, Ding, Gao, Ge, et~al.]{qwen3technicalreport}
Shuai Bai, Yuxuan Cai, Ruizhe Chen, Keqin Chen, Xionghui Chen, Zesen Cheng, Lianghao Deng, Wei Ding, Chang Gao, Chunjiang Ge, et~al.
\newblock Qwen3-vl technical report.
\newblock \emph{arXiv preprint arXiv:2511.21631}, 2025{\natexlab{a}}.

\bibitem[Bai et~al.(2025{\natexlab{b}})Bai, Chen, Liu, Wang, Ge, Song, Dang, Wang, Wang, Tang, Zhong, Zhu, Yang, Li, Wan, Wang, Ding, Fu, Xu, Ye, Zhang, Xie, Cheng, Zhang, Yang, Xu, and Lin]{qwenvl}
Shuai Bai, Keqin Chen, Xuejing Liu, Jialin Wang, Wenbin Ge, Sibo Song, Kai Dang, Peng Wang, Shijie Wang, Jun Tang, Humen Zhong, Yuanzhi Zhu, Mingkun Yang, Zhaohai Li, Jianqiang Wan, Pengfei Wang, Wei Ding, Zheren Fu, Yiheng Xu, Jiabo Ye, Xi Zhang, Tianbao Xie, Zesen Cheng, Hang Zhang, Zhibo Yang, Haiyang Xu, and Junyang Lin.
\newblock Qwen2.5-vl technical report.
\newblock \emph{arXiv preprint arXiv:2502.13923}, 2025{\natexlab{b}}.

\bibitem[Beaumont and Schuhmann(2022)]{laionaes}
Romain Beaumont and Christoph Schuhmann.
\newblock Laion-aesthetics predictor.
\newblock \url{https://github.com/LAION-AI/aesthetic-predictor}, 2022.

\bibitem[Cai et~al.(2023)Cai, Ma, Wang, and Li]{cai2023image}
Qiang Cai, Mengxu Ma, Chen Wang, and Haisheng Li.
\newblock Image neural style transfer: A review.
\newblock \emph{Computers and Electrical Engineering}, 108:\penalty0 108723, 2023.

\bibitem[Chung et~al.(2024{\natexlab{a}})Chung, Hyun, and Heo]{Chung_2024_CVPR}
Jiwoo Chung, Sangeek Hyun, and Jae-Pil Heo.
\newblock Style injection in diffusion: A training-free approach for adapting large-scale diffusion models for style transfer.
\newblock In \emph{Proceedings of the IEEE/CVF Conference on Computer Vision and Pattern Recognition (CVPR)}, pages 8795--8805, 2024{\natexlab{a}}.

\bibitem[Chung et~al.(2024{\natexlab{b}})Chung, Hyun, and Heo]{chung2024style}
Jiwoo Chung, Sangeek Hyun, and Jae-Pil Heo.
\newblock Style injection in diffusion: A training-free approach for adapting large-scale diffusion models for style transfer.
\newblock In \emph{Proceedings of the IEEE/CVF conference on computer vision and pattern recognition}, pages 8795--8805, 2024{\natexlab{b}}.

\bibitem[Dai et~al.(2024)Dai, Deng, Zhao, Xu, Gao, Chen, Li, Zeng, Yu, Wu, et~al.]{dai2024deepseekmoe}
Damai Dai, Chengqi Deng, Chenggang Zhao, RX Xu, Huazuo Gao, Deli Chen, Jiashi Li, Wangding Zeng, Xingkai Yu, Yu Wu, et~al.
\newblock Deepseekmoe: Towards ultimate expert specialization in mixture-of-experts language models.
\newblock \emph{arXiv preprint arXiv:2401.06066}, 2024.

\bibitem[Esser et~al.(2024)Esser, Kulal, Blattmann, Entezari, M{\"u}ller, Saini, Levi, Lorenz, Sauer, Boesel, et~al.]{esser2024scaling}
Patrick Esser, Sumith Kulal, Andreas Blattmann, Rahim Entezari, Jonas M{\"u}ller, Harry Saini, Yam Levi, Dominik Lorenz, Axel Sauer, Frederic Boesel, et~al.
\newblock Scaling rectified flow transformers for high-resolution image synthesis.
\newblock In \emph{Forty-first international conference on machine learning}, 2024.

\bibitem[Fedus et~al.(2022)Fedus, Zoph, and Shazeer]{fedus2022switch}
William Fedus, Barret Zoph, and Noam Shazeer.
\newblock Switch transformers: Scaling to trillion parameter models with simple and efficient sparsity.
\newblock \emph{Journal of Machine Learning Research}, 23\penalty0 (120):\penalty0 1--39, 2022.

\bibitem[Fortin et~al.(2025)Fortin, Vernade, Kampf, and Reshi]{nanobanana}
Alisa Fortin, Guillaume Vernade, Kat Kampf, and Ammaar Reshi.
\newblock Introducing gemini 2.5 flash image, our state-of-the-art image model.
\newblock https://developers.googleblog.com/en/introducing-gemini-2-5-flash-image/, 2025.

\bibitem[Frenkel et~al.(2024)Frenkel, Vinker, Shamir, and Cohen-Or]{frenkel2024implicit}
Yarden Frenkel, Yael Vinker, Ariel Shamir, and Daniel Cohen-Or.
\newblock Implicit style-content separation using b-lora.
\newblock In \emph{European Conference on Computer Vision}, pages 181--198. Springer, 2024.

\bibitem[Fu et~al.(2023)Fu, Tamir, Sundaram, Chai, Zhang, Dekel, and Isola]{fu2023dreamsim}
Stephanie Fu, Netanel Tamir, Shobhita Sundaram, Lucy Chai, Richard Zhang, Tali Dekel, and Phillip Isola.
\newblock Dreamsim: Learning new dimensions of human visual similarity using synthetic data, 2023.

\bibitem[Gao et~al.(2025)Gao, Sun, Liu, Tang, Zeng, Qi, Chen, and Zhao]{gao2025styleshot}
Junyao Gao, Yanan Sun, Yanchen Liu, Yinhao Tang, Yanhong Zeng, Ding Qi, Kai Chen, and Cairong Zhao.
\newblock Styleshot: A snapshot on any style.
\newblock \emph{IEEE Transactions on Pattern Analysis and Machine Intelligence}, 2025.

\bibitem[Gatys et~al.(2016)Gatys, Ecker, and Bethge]{gatys2016image}
Leon~A Gatys, Alexander~S Ecker, and Matthias Bethge.
\newblock Image style transfer using convolutional neural networks.
\newblock In \emph{Proceedings of the IEEE conference on computer vision and pattern recognition}, pages 2414--2423, 2016.

\bibitem[Gatys et~al.(2017)Gatys, Ecker, Bethge, Hertzmann, and Shechtman]{gatys2017controlling}
Leon~A Gatys, Alexander~S Ecker, Matthias Bethge, Aaron Hertzmann, and Eli Shechtman.
\newblock Controlling perceptual factors in neural style transfer.
\newblock In \emph{Proceedings of the IEEE conference on computer vision and pattern recognition}, pages 3985--3993, 2017.

\bibitem[Hu et~al.(2022)Hu, Shen, Wallis, Allen-Zhu, Li, Wang, Wang, Chen, et~al.]{hu2022lora}
Edward~J Hu, Yelong Shen, Phillip Wallis, Zeyuan Allen-Zhu, Yuanzhi Li, Shean Wang, Lu Wang, Weizhu Chen, et~al.
\newblock Lora: Low-rank adaptation of large language models.
\newblock \emph{ICLR}, 1\penalty0 (2):\penalty0 3, 2022.

\bibitem[Jacobs et~al.(1991)Jacobs, Jordan, Nowlan, and Hinton]{Jacobs1991moe}
Robert~A. Jacobs, Michael~I. Jordan, Steven~J. Nowlan, and Geoffrey~E. Hinton.
\newblock Adaptive mixtures of local experts.
\newblock \emph{Neural Computation}, 3\penalty0 (1):\penalty0 79--87, 1991.

\bibitem[Jeong et~al.(2024)Jeong, Kwon, and Uh]{jeong2024training}
Jaeseok Jeong, Mingi Kwon, and Youngjung Uh.
\newblock Training-free content injection using h-space in diffusion models.
\newblock In \emph{Proceedings of the IEEE/CVF Winter Conference on Applications of Computer Vision}, pages 5151--5161, 2024.

\bibitem[Jing et~al.(2019)Jing, Yang, Feng, Ye, Yu, and Song]{jing2019neural}
Yongcheng Jing, Yezhou Yang, Zunlei Feng, Jingwen Ye, Yizhou Yu, and Mingli Song.
\newblock Neural style transfer: A review.
\newblock \emph{IEEE transactions on visualization and computer graphics}, 26\penalty0 (11):\penalty0 3365--3385, 2019.

\bibitem[Jordan and Jacobs(1994)]{jordan1994hierarchical}
Michael~I Jordan and Robert~A Jacobs.
\newblock Hierarchical mixtures of experts and the em algorithm.
\newblock \emph{Neural computation}, 6\penalty0 (2):\penalty0 181--214, 1994.

\bibitem[Kingma and Welling(2013)]{kingma2013auto}
Diederik~P Kingma and Max Welling.
\newblock Auto-encoding variational bayes.
\newblock \emph{arXiv preprint arXiv:1312.6114}, 2013.

\bibitem[Kolkin et~al.(2019)Kolkin, Salavon, and Shakhnarovich]{kolkin2019style}
Nicholas Kolkin, Jason Salavon, and Gregory Shakhnarovich.
\newblock Style transfer by relaxed optimal transport and self-similarity.
\newblock In \emph{Proceedings of the IEEE/CVF conference on computer vision and pattern recognition}, pages 10051--10060, 2019.

\bibitem[Labs(2024)]{flux2024}
Black~Forest Labs.
\newblock Flux.
\newblock \url{https://github.com/black-forest-labs/flux}, 2024.

\bibitem[Labs et~al.(2025)Labs, Batifol, Blattmann, Boesel, Consul, Diagne, Dockhorn, English, English, Esser, Kulal, Lacey, Levi, Li, Lorenz, Müller, Podell, Rombach, Saini, Sauer, and Smith]{labs2025flux1kontextflowmatching}
Black~Forest Labs, Stephen Batifol, Andreas Blattmann, Frederic Boesel, Saksham Consul, Cyril Diagne, Tim Dockhorn, Jack English, Zion English, Patrick Esser, Sumith Kulal, Kyle Lacey, Yam Levi, Cheng Li, Dominik Lorenz, Jonas Müller, Dustin Podell, Robin Rombach, Harry Saini, Axel Sauer, and Luke Smith.
\newblock Flux.1 kontext: Flow matching for in-context image generation and editing in latent space, 2025.

\bibitem[Lee et~al.(2024)Lee, Kim, Byeon, Oh, In, Park, Yoon, Hong, Kim, and Kim]{lee2024audio}
Seung~Hyun Lee, Sieun Kim, Wonmin Byeon, Gyeongrok Oh, Sumin In, Hyeongcheol Park, Sang~Ho Yoon, Sung-Hee Hong, Jinkyu Kim, and Sangpil Kim.
\newblock Audio-guided implicit neural representation for local image stylization.
\newblock \emph{Computational Visual Media}, 10\penalty0 (6):\penalty0 1185--1204, 2024.

\bibitem[Li et~al.(2024)Li, Fang, Zou, Gong, Zheng, Wang, Chen, and Yang]{li2024styletokenizer}
Wen Li, Muyuan Fang, Cheng Zou, Biao Gong, Ruobing Zheng, Meng Wang, Jingdong Chen, and Ming Yang.
\newblock Styletokenizer: Defining image style by a single instance for controlling diffusion models.
\newblock In \emph{European Conference on Computer Vision}, pages 110--126. Springer, 2024.

\bibitem[Liu et~al.(2025)Liu, Zheng, and Yang]{liu2025sad}
Yilong Liu, Houwen Zheng, and Shuojin Yang.
\newblock Sad: Style-aware diffusion adaptation for few-shot style transfer image generation.
\newblock \emph{Computational Visual Media}, 2025.

\bibitem[Mou et~al.(2025)Mou, Wu, Wu, Guo, Zhang, Cheng, Luo, Ding, Zhang, Li, et~al.]{mou2025dreamo}
Chong Mou, Yanze Wu, Wenxu Wu, Zinan Guo, Pengze Zhang, Yufeng Cheng, Yiming Luo, Fei Ding, Shiwen Zhang, Xinghui Li, et~al.
\newblock Dreamo: A unified framework for image customization.
\newblock \emph{arXiv preprint arXiv:2504.16915}, 2025.

\bibitem[Oord et~al.(2018)Oord, Li, and Vinyals]{oord2018representation}
Aaron van~den Oord, Yazhe Li, and Oriol Vinyals.
\newblock Representation learning with contrastive predictive coding.
\newblock \emph{arXiv preprint arXiv:1807.03748}, 2018.

\bibitem[Ouyang et~al.(2025)Ouyang, Li, and Hou]{ouyang2025k}
Ziheng Ouyang, Zhen Li, and Qibin Hou.
\newblock K-lora: Unlocking training-free fusion of any subject and style loras.
\newblock In \emph{CVPR}, 2025.

\bibitem[Peebles and Xie(2023)]{peebles2023scalable}
William Peebles and Saining Xie.
\newblock Scalable diffusion models with transformers.
\newblock In \emph{Proceedings of the IEEE/CVF international conference on computer vision}, pages 4195--4205, 2023.

\bibitem[Podell et~al.(2023)Podell, English, Lacey, Blattmann, Dockhorn, M{\"u}ller, Penna, and Rombach]{podell2023sdxl}
Dustin Podell, Zion English, Kyle Lacey, Andreas Blattmann, Tim Dockhorn, Jonas M{\"u}ller, Joe Penna, and Robin Rombach.
\newblock Sdxl: Improving latent diffusion models for high-resolution image synthesis.
\newblock \emph{arXiv preprint arXiv:2307.01952}, 2023.

\bibitem[Qin et~al.(2025)Qin, Li, Gomez-Villa, Yang, Wang, Wang, and van~de Weijer]{qin2025free}
Jiang Qin, Senmao Li, Alexandra Gomez-Villa, Shiqi Yang, Yaxing Wang, Kai Wang, and Joost van~de Weijer.
\newblock Free-lunch color-texture disentanglement for stylized image generation.
\newblock \emph{arXiv preprint arXiv:2503.14275}, 2025.

\bibitem[Radford et~al.(2021)Radford, Kim, Hallacy, Ramesh, Goh, Agarwal, Sastry, Askell, Mishkin, Clark, et~al.]{radford2021learning}
Alec Radford, Jong~Wook Kim, Chris Hallacy, Aditya Ramesh, Gabriel Goh, Sandhini Agarwal, Girish Sastry, Amanda Askell, Pamela Mishkin, Jack Clark, et~al.
\newblock Learning transferable visual models from natural language supervision.
\newblock In \emph{International conference on machine learning}, pages 8748--8763. PmLR, 2021.

\bibitem[Schuhmann et~al.(2022)Schuhmann, Beaumont, Vencu, Gordon, Wightman, Cherti, Coombes, Katta, Mullis, Wortsman, et~al.]{schuhmann2022laion}
Christoph Schuhmann, Romain Beaumont, Richard Vencu, Cade Gordon, Ross Wightman, Mehdi Cherti, Theo Coombes, Aarush Katta, Clayton Mullis, Mitchell Wortsman, et~al.
\newblock Laion-5b: An open large-scale dataset for training next generation image-text models.
\newblock \emph{Advances in neural information processing systems}, 35:\penalty0 25278--25294, 2022.

\bibitem[Shazeer et~al.(2017)Shazeer, Mirhoseini, Maziarz, Davis, Le, Hinton, and Dean]{shazeer2017outrageously}
Noam Shazeer, Azalia Mirhoseini, Krzysztof Maziarz, Andy Davis, Quoc Le, Geoffrey Hinton, and Jeff Dean.
\newblock Outrageously large neural networks: The sparsely-gated mixture-of-experts layer.
\newblock \emph{arXiv preprint arXiv:1701.06538}, 2017.

\bibitem[Sohn et~al.(2023)Sohn, Ruiz, Lee, Chin, Blok, Chang, Barber, Jiang, Entis, Li, et~al.]{sohn2023styledrop}
Kihyuk Sohn, Nataniel Ruiz, Kimin Lee, Daniel~Castro Chin, Irina Blok, Huiwen Chang, Jarred Barber, Lu Jiang, Glenn Entis, Yuanzhen Li, et~al.
\newblock Styledrop: Text-to-image generation in any style.
\newblock \emph{arXiv preprint arXiv:2306.00983}, 2023.

\bibitem[Somepalli et~al.(2024)Somepalli, Gupta, Gupta, Palta, Goldblum, Geiping, Shrivastava, and Goldstein]{somepalli2024measuring}
Gowthami Somepalli, Anubhav Gupta, Kamal Gupta, Shramay Palta, Micah Goldblum, Jonas Geiping, Abhinav Shrivastava, and Tom Goldstein.
\newblock Measuring style similarity in diffusion models.
\newblock \emph{arXiv preprint arXiv:2404.01292}, 2024.

\bibitem[Song et~al.(2025)Song, Liu, and Shou]{song2025omniconsistency}
Yiren Song, Cheng Liu, and Mike~Zheng Shou.
\newblock Omniconsistency: Learning style-agnostic consistency from paired stylization data.
\newblock \emph{arXiv preprint arXiv:2505.18445}, 2025.

\bibitem[Wang et~al.(2024{\natexlab{a}})Wang, Spinelli, Wang, Bai, Qin, and Chen]{wang2024instantstyle}
Haofan Wang, Matteo Spinelli, Qixun Wang, Xu Bai, Zekui Qin, and Anthony Chen.
\newblock Instantstyle: Free lunch towards style-preserving in text-to-image generation.
\newblock \emph{arXiv preprint arXiv:2404.02733}, 2024{\natexlab{a}}.

\bibitem[Wang et~al.(2024{\natexlab{b}})Wang, Xing, Huang, Ai, Wang, and Bai]{wang2024plus}
Haofan Wang, Peng Xing, Renyuan Huang, Hao Ai, Qixun Wang, and Xu Bai.
\newblock Instantstyle-plus: Style transfer with content-preserving in text-to-image generation.
\newblock \emph{arXiv preprint arXiv:2407.00788}, 2024{\natexlab{b}}.

\bibitem[Wang et~al.(2025{\natexlab{a}})Wang, Liu, Lin, Liu, Yi, Wang, and Ma]{wang2025omnistyle}
Ye Wang, Ruiqi Liu, Jiang Lin, Fei Liu, Zili Yi, Yilin Wang, and Rui Ma.
\newblock Omnistyle: Filtering high quality style transfer data at scale.
\newblock In \emph{Proceedings of the Computer Vision and Pattern Recognition Conference}, pages 7847--7856, 2025{\natexlab{a}}.

\bibitem[Wang et~al.(2025{\natexlab{b}})Wang, Yi, Zhang, Zheng, Xie, Lin, Wang, and Ma]{wang2025omnistyle2}
Ye Wang, Zili Yi, Yibo Zhang, Peng Zheng, Xuping Xie, Jiang Lin, Yilin Wang, and Rui Ma.
\newblock Omnistyle2: Scalable and high quality artistic style transfer data generation via destylization.
\newblock \emph{arXiv preprint arXiv:2509.05970}, 2025{\natexlab{b}}.

\bibitem[Wu et~al.(2025{\natexlab{a}})Wu, Li, Zhou, Lin, Gao, Yan, ming Yin, Bai, Xu, Chen, Chen, Tang, Zhang, Wang, Yang, Yu, Cheng, Liu, Li, Zhang, Meng, Wei, Ni, Chen, Cao, Peng, Qu, Wu, Wang, Yu, Wen, Feng, Xu, Wang, Zhang, Zhu, Wu, Cai, and Liu]{wu2025qwenimagetechnicalreport}
Chenfei Wu, Jiahao Li, Jingren Zhou, Junyang Lin, Kaiyuan Gao, Kun Yan, Sheng ming Yin, Shuai Bai, Xiao Xu, Yilei Chen, Yuxiang Chen, Zecheng Tang, Zekai Zhang, Zhengyi Wang, An Yang, Bowen Yu, Chen Cheng, Dayiheng Liu, Deqing Li, Hang Zhang, Hao Meng, Hu Wei, Jingyuan Ni, Kai Chen, Kuan Cao, Liang Peng, Lin Qu, Minggang Wu, Peng Wang, Shuting Yu, Tingkun Wen, Wensen Feng, Xiaoxiao Xu, Yi Wang, Yichang Zhang, Yongqiang Zhu, Yujia Wu, Yuxuan Cai, and Zenan Liu.
\newblock Qwen-image technical report, 2025{\natexlab{a}}.

\bibitem[Wu et~al.(2025{\natexlab{b}})Wu, Zheng, Yan, Xiao, Luo, Wang, Li, Jiang, Liu, Zhou, et~al.]{wu2025omnigen2}
Chenyuan Wu, Pengfei Zheng, Ruiran Yan, Shitao Xiao, Xin Luo, Yueze Wang, Wanli Li, Xiyan Jiang, Yexin Liu, Junjie Zhou, et~al.
\newblock Omnigen2: Exploration to advanced multimodal generation.
\newblock \emph{arXiv preprint arXiv:2506.18871}, 2025{\natexlab{b}}.

\bibitem[Wu et~al.(2025{\natexlab{c}})Wu, Huang, Cheng, Wu, Tian, Luo, Ding, and He]{wu2025uso}
Shaojin Wu, Mengqi Huang, Yufeng Cheng, Wenxu Wu, Jiahe Tian, Yiming Luo, Fei Ding, and Qian He.
\newblock Uso: Unified style and subject-driven generation via disentangled and reward learning.
\newblock \emph{arXiv preprint arXiv:2508.18966}, 2025{\natexlab{c}}.

\bibitem[Wu et~al.(2025{\natexlab{d}})Wu, Jiang, Lu, Wang, Huang, Qin, and Li]{wu2025multicrafter}
Tao Wu, Yibo Jiang, Yehao Lu, Zhizhong Wang, Zeyi Huang, Zequn Qin, and Xi Li.
\newblock Multicrafter: High-fidelity multi-subject generation via spatially disentangled attention and identity-aware reinforcement learning.
\newblock \emph{arXiv preprint arXiv:2509.21953}, 2025{\natexlab{d}}.

\bibitem[Wu et~al.(2020)Wu, Song, Zhou, Gong, and Huang]{wu2020efanet}
Zhijie Wu, Chunjin Song, Yang Zhou, Minglun Gong, and Hui Huang.
\newblock Efanet: Exchangeable feature alignment network for arbitrary style transfer.
\newblock In \emph{Proceedings of the AAAI Conference on Artificial Intelligence}, pages 12305--12312, 2020.

\bibitem[Wu et~al.(2022)Wu, Song, Chen, Guo, and Huang]{wu2022completeness}
Zhijie Wu, Chunjin Song, Guanxiong Chen, Sheng Guo, and Weilin Huang.
\newblock Completeness and coherence learning for fast arbitrary style transfer.
\newblock \emph{Transactions on Machine Learning Research}, 2022.

\bibitem[Xing et~al.(2024)Xing, Wang, Sun, Wang, Bai, Ai, Huang, and Li]{xing2024csgo}
Peng Xing, Haofan Wang, Yanpeng Sun, Qixun Wang, Xu Bai, Hao Ai, Renyuan Huang, and Zechao Li.
\newblock Csgo: Content-style composition in text-to-image generation.
\newblock \emph{arXiv preprint arXiv:2408.16766}, 2024.

\bibitem[Xu et~al.(2024)Xu, Wang, Xiao, Liu, and Chen]{xu2024freetuner}
Youcan Xu, Zhen Wang, Jun Xiao, Wei Liu, and Long Chen.
\newblock Freetuner: Any subject in any style with training-free diffusion.
\newblock \emph{arXiv preprint arXiv:2405.14201}, 2024.

\bibitem[Yang et~al.(2024)Yang, Jiang, Hou, Zhang, Chen, and Li]{yang2024multi}
Yuqi Yang, Peng-Tao Jiang, Qibin Hou, Hao Zhang, Jinwei Chen, and Bo Li.
\newblock Multi-task dense prediction via mixture of low-rank experts.
\newblock In \emph{Proceedings of the IEEE/CVF conference on computer vision and pattern recognition}, pages 27927--27937, 2024.

\bibitem[Zhai et~al.(2023)Zhai, Mustafa, Kolesnikov, and Beyer]{zhai2023sigmoid}
Xiaohua Zhai, Basil Mustafa, Alexander Kolesnikov, and Lucas Beyer.
\newblock Sigmoid loss for language image pre-training, 2023.

\bibitem[Zhang et~al.(2022)Zhang, Li, Liu, Zhang, Su, Zhu, Ni, and Shum]{zhang2022dino}
Hao Zhang, Feng Li, Shilong Liu, Lei Zhang, Hang Su, Jun Zhu, Lionel~M. Ni, and Heung-Yeung Shum.
\newblock Dino: Detr with improved denoising anchor boxes for end-to-end object detection, 2022.

\bibitem[Zhang et~al.(2023{\natexlab{a}})Zhang, Rao, and Agrawala]{zhang2023adding}
Lvmin Zhang, Anyi Rao, and Maneesh Agrawala.
\newblock Adding conditional control to text-to-image diffusion models, 2023{\natexlab{a}}.

\bibitem[Zhang et~al.(2023{\natexlab{b}})Zhang, Huang, Tang, Huang, Ma, Dong, and Xu]{zhang2023inversion}
Yuxin Zhang, Nisha Huang, Fan Tang, Haibin Huang, Chongyang Ma, Weiming Dong, and Changsheng Xu.
\newblock Inversion-based style transfer with diffusion models.
\newblock In \emph{Proceedings of the IEEE/CVF conference on computer vision and pattern recognition}, pages 10146--10156, 2023{\natexlab{b}}.

\bibitem[Zhang et~al.(2024)Zhang, Song, Liu, Wang, Yu, Tang, Li, Tang, Hu, Pan, et~al.]{zhang2024ssr}
Yuxuan Zhang, Yiren Song, Jiaming Liu, Rui Wang, Jinpeng Yu, Hao Tang, Huaxia Li, Xu Tang, Yao Hu, Han Pan, et~al.
\newblock Ssr-encoder: Encoding selective subject representation for subject-driven generation.
\newblock In \emph{Proceedings of the IEEE/CVF Conference on Computer Vision and Pattern Recognition}, pages 8069--8078, 2024.

\bibitem[Zhang et~al.(2025{\natexlab{a}})Zhang, Yuan, Song, Wang, and Liu]{zhang2025easycontrol}
Yuxuan Zhang, Yirui Yuan, Yiren Song, Haofan Wang, and Jiaming Liu.
\newblock Easycontrol: Adding efficient and flexible control for diffusion transformer.
\newblock \emph{arXiv preprint arXiv:2503.07027}, 2025{\natexlab{a}}.

\bibitem[Zhang et~al.(2025{\natexlab{b}})Zhang, Xie, Lu, Yang, and Yang]{zhang2025context}
Zechuan Zhang, Ji Xie, Yu Lu, Zongxin Yang, and Yi Yang.
\newblock In-context edit: Enabling instructional image editing with in-context generation in large scale diffusion transformer.
\newblock \emph{arXiv preprint arXiv:2504.20690}, 2025{\natexlab{b}}.

\bibitem[Zhang et~al.(2025{\natexlab{c}})Zhang, Xie, Lu, Yang, and Yang]{zhangenabling}
Zechuan Zhang, Ji Xie, Yu Lu, Zongxin Yang, and Yi Yang.
\newblock Enabling instructional image editing with in-context generation in large scale diffusion transformer.
\newblock In \emph{The Thirty-ninth Annual Conference on Neural Information Processing Systems}, 2025{\natexlab{c}}.

\bibitem[Zhou et~al.(2025)Zhou, Gao, Chen, and Huang]{zhou2025attention}
Yang Zhou, Xu Gao, Zichong Chen, and Hui Huang.
\newblock Attention distillation: A unified approach to visual characteristics transfer.
\newblock In \emph{Proceedings of the Computer Vision and Pattern Recognition Conference}, pages 18270--18280, 2025.

\end{thebibliography}
}

\newpage
\appendix
\twocolumn[{%
\begin{center}
    {\LARGE \textbf{Supplementary Materials}}\\[0.5em]
\end{center}

}]

\noindent Our supplementary material is organized as follows:
\begin{enumerate}
    \item We detail our evaluation metrics, including the proposed Qwen Semantic Score (Sec.~\ref{sec:qwen_score}).
    \item We present additional implementation details for our experiments (Sec.~\ref{sec:experiments_details}).
    \item We show the prompts used during our dataset construction and their corresponding results (Sec.~\ref{sec:dataset_details}).
    \item We showcase example comparisons between our dataset and other existing datasets (Sec.~\ref{sec:dataset_comparison}).
    \item We provide expanded comparisons of our method against additional competing methods and on more unseen styles (Sec.~\ref{sec:expanded_comparisons}).
    \item We show more generation results from our method (Sec.~\ref{sec:more_results}).
\end{enumerate}

\section{Detailed Evaluation Metrics}
\label{sec:qwen_score}

The Qwen Semantic Score is proposed to evaluate whether a stylization process utilizes semantic information from the style reference image.
To assess this, we feed the style reference image and the stylized output image into a VLM, tasking it with determining if semantic information from the style image was incorporated. 
The specific prompt used for this VLM-based evaluation is detailed in Listing~\ref{lst:qwen_prompt}.

As shown in the prompt, \texttt{style\_path} corresponds to the style reference image and \texttt{output\_path} corresponds to the stylized output image. We use Qwen3-VL-8B-Instruct \cite{qwen3technicalreport} as the VLM.
If the model's response begins with 'YES', the score is 1; otherwise, it is 0.

In practice, for the \emph{Stylized Dataset Curation} section, we evaluate the semantic focus of each style by randomly sampling five content-style-stylized triplets.
We compute the Qwen Semantic Score for each triplet and sum them to produce a final semantic focus score for that style, ranging from 0 to 5.
Conversely, in the \emph{Quantitative Comparisons} section, we evaluate a model's ability to transfer semantic information. 
We compute the Qwen Semantic Score for every stylized triplet in the test set and then average these scores.
The resulting metric is a value between 0 and 1.

Additionally, the reported CLIP score averages the similarities to both the content and style references, preventing models from merely copying the content image.

\section{Experiments Details}
\label{sec:experiments_details}
\subsection{IoU Experiment Details}
In the IoU experiment presented in the \textit{Discussion} section, we calculate the IoU of the expert indices selected for similar and dissimilar styles.
Specifically, we ensure all reference images share the same content.
First, we randomly select an anchor image $I$ from our test set.
We then define the candidate pool $D$ as the set of all other test images that share the same content as $I$.
We compute the CLIP similarity between $I$ and all $I_j \in D$, then designate the similar style $I_s$ and dissimilar style $I_d$ as:
\begin{equation} \label{eq:is}
I_s = \arg\max_{I_j \in D} \text{CLIPSim}(I, I_j),
\end{equation}
\begin{equation} \label{eq:id}
I_d = \arg\min_{I_j \in D} \text{CLIPSim}(I, I_j).
\end{equation}

For each MoE-enabled layer $l$, let $E_l(I)$ be the set of expert indices selected by the gating network $G_l$ for image $I$.
We then compute the IoU for the similar pair $(I, I_s)$ and the dissimilar pair $(I, I_d)$ at this layer:
\begin{equation} \label{eq:iou_sim}
\text{IoU}_{\text{sim}, l} = \frac{|E_l(I) \cap E_l(I_s)|}{|E_l(I) \cup E_l(I_s)|},
\end{equation}
\begin{equation} \label{eq:iou_dissim}
\text{IoU}_{\text{dissim}, l} = \frac{|E_l(I) \cap E_l(I_d)|}{|E_l(I) \cup E_l(I_d)|}.
\end{equation}
We repeat this process for 100 sampled triplets $(I, I_s, I_d)$ and compute the average IoU for each layer.
In the table, “Early Stage," “Mid Stage," and “Late Stage" correspond to the first 1/3, middle 1/3, and final 1/3 of the MoE-injected layers, respectively.

\subsection{Evaluation Prompts}

Some stylization methods require text prompts to function. 
Methods like OmniStyle \cite{wang2025omnistyle} and CSGO \cite{xing2024csgo} require a \emph{content prompt} to describe the subject. 
Others, such as DreamO \cite{mou2025dreamo}, OmniGen2 \cite{wu2025omnigen2}, and Qwen-Image-Edit \cite{wu2025qwenimagetechnicalreport}, treat stylization as a sub-task and require a \emph{task prompt} to activate it.
For the first category, we supply a clean content prompt during evaluation to ensure optimal performance.
For the second, we identified a robust, general-purpose task prompt through testing: \texttt{"Make the whole first image have the same image style as the second image."}
We use this prompt in all evaluations for these methods.

\begin{figure*}
    \centering
    \includegraphics[width=1\linewidth]{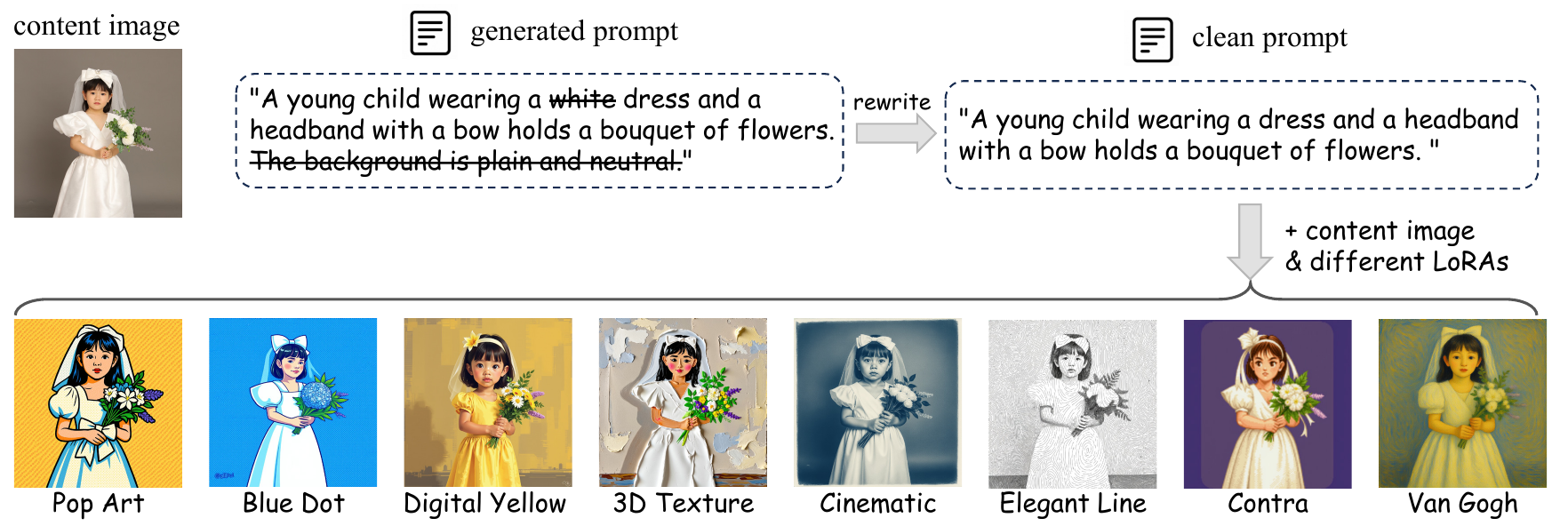}
    \caption{Illustration of our prompt filtering process. Initial VL-generated captions often retain descriptors (\eg, colors, atmosphere) that conflict with styles defined by their own intrinsic properties. This is detrimental to styles with unique color palettes (\eg, blue dot, Digital Yellow, Van Gogh, Elegant Line) or distinct atmospheres (\eg, Pop Art, 3D Texture, Cinematic, Contra). Our rewriting step removes these conflicting descriptors to produce clean prompts, ensuring consistent stylization.}
    \label{fig:filter_prompt}
\end{figure*}

\section{Dataset Curation Details}
\label{sec:dataset_details}
To generate a dataset with high style consistency, we must obtain clean prompts that are free of any style descriptors, as these descriptors could adversely affect the stylization process.
To this end, we first generate an initial caption by feeding the content image and the prompt from Listing~\ref{lst:gen_caption} into Qwen3-VL \cite{qwen3technicalreport}.
Although Listing~\ref{lst:gen_caption} explicitly requests a content-only description, the resulting captions often retain style information.
For example, as shown in \figref{fig:filter_prompt}, an initial caption might be: ``A young child wearing a white dress... The background is plain and neutral.''
These descriptors are detrimental because many artistic styles define their own intrinsic color palettes or background compositions. A prompt specifying ``white'' or a ``plain'' background would conflict with such styles and degrade the stylization quality.
Therefore, we perform an additional rewriting step. We use Qwen3 \cite{qwen3technicalreport} and the prompt from Listing~\ref{lst:rewrite_caption} to remove this specific style information, yielding a final clean caption such as: ``A young child wearing a dress and a headband with a bow holds a bouquet of flowers.''

To filter out triplets where the layout has changed significantly, we primarily check for variations in composition and specific attributes, such as the number or gender of persons.
The prompts used for this filtering process are provided in Listing~\ref{lst:filter1} and Listing~\ref{lst:filter2}.

\section{Dataset Comparison}
\label{sec:dataset_comparison}
\begin{figure*}
    \centering
    \includegraphics[width=1\linewidth]{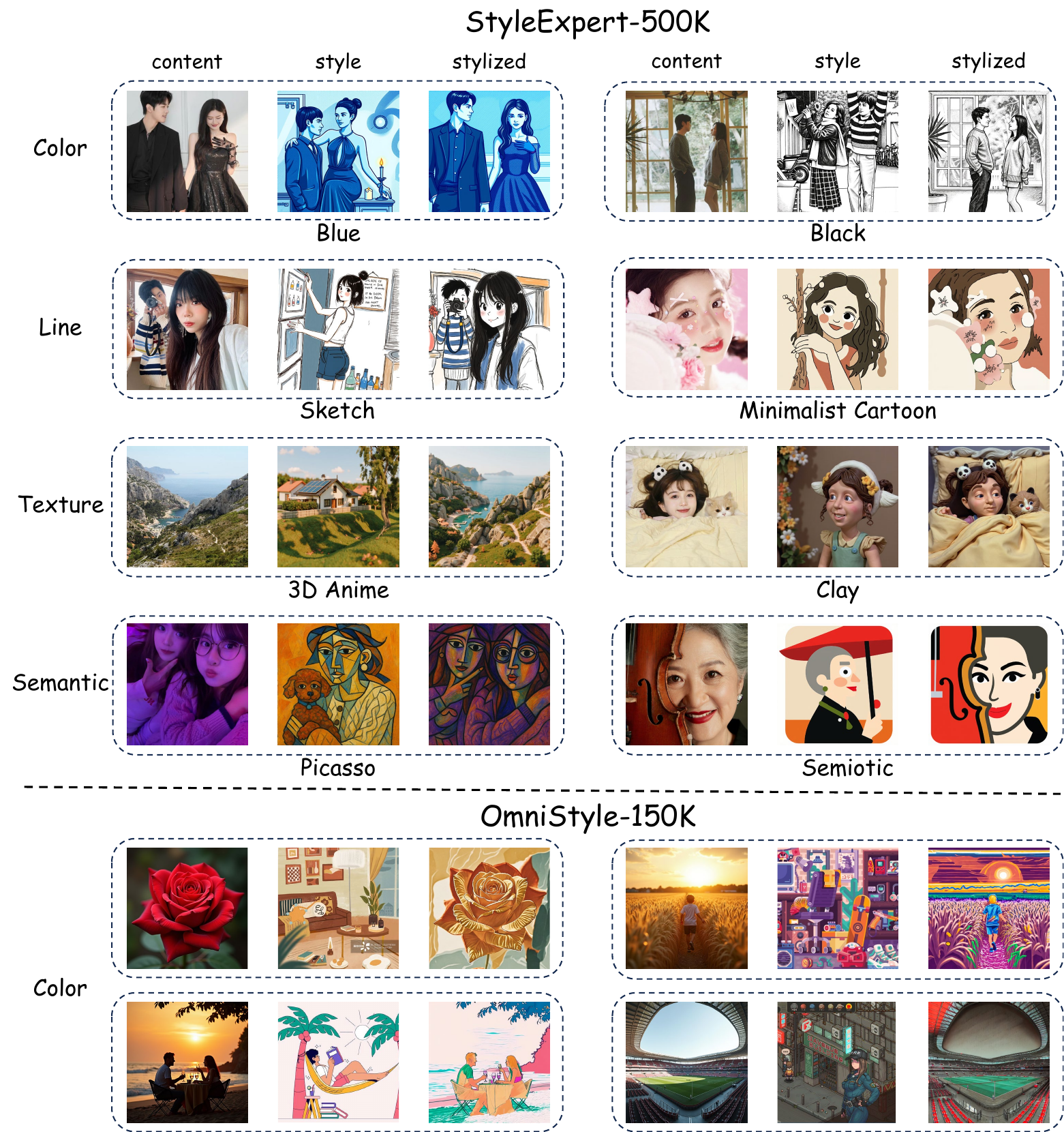}
    \caption{Comparison of data samples from our \ourMthd{}-500K dataset and the OmniStyle-150K dataset.}
    \label{fig:dataset_comparison}
\end{figure*}

To further illustrate the differences between our dataset and previous datasets, \figref{fig:dataset_comparison} compares our \ourMthd{}-500K dataset with the previous OmniStyle-150K dataset.
Our dataset exhibits significant stylistic diversity, with styles categorized by their depth of semantic focus, spanning Color, Line, Texture, and Semantic levels.
In contrast, the majority of examples in OmniStyle-150K degenerate to simple color transfer, failing to utilize deeper semantic information from the style reference.
For instance, they fail to capture the "illustration style" (Row 1, Col 1) or the "symbolic style" (Row 1, Col 2).
Furthermore, because the reference colors are not always suitable for the content image, these stylizations often exhibit poor aesthetic quality.

Beyond qualitative observations, \tabref{tab:dataset_comparison} provides a quantitative comparison between the two datasets. 
Unlike OmniStyle-150K, our \ourMthd{}-500K successfully maintains color-semantic balance. 
Furthermore, it outperforms the baseline across most metrics, particularly achieving a significantly higher CSD score and a lower DreamSim distance. 
While performing comparably on DINO, our dataset yields better CLIP similarity and Aesthetic scores, quantitatively demonstrating its superior stylistic diversity and overall visual quality.

\begin{table}[h!]

\centering
\scriptsize
\caption{Quantitative comparison between OmniStyle-150K and our \ourMthd{}-500K.}
\vspace{-10pt}
\setlength{\tabcolsep}{2pt}
\begin{tabular}{@{}cccccccc@{}}
\toprule
          &    Color-Semantic     & CLIP & DINO & CSD  & Aesthetic & DreamSim \\ 
Dataset     &  Balance & ($\uparrow$) & ($\uparrow$) & ($\uparrow$)  & ($\uparrow$) & ($\downarrow$) \\ \hline

{OmniStyle-150K} & \ding{55} & {67.68} & \textbf{67.71}          & 38.47 & {6.08}        & {59.49}       \\
{StyleExpert-500K} & \checkmark  & \textbf{68.41} & 66.75          & \textbf{74.49} & \textbf{6.56 }      & \textbf{36.19}       \\ 
\bottomrule
\end{tabular}
\vspace{-12pt}
\label{tab:dataset_comparison}
\end{table}

\section{User Study}
To subjectively assess our method, we conducted a user study involving 30 participants and 1,200 total votes. Users were asked to select the best stylized output from randomized baseline results, evaluating three main aspects: style consistency, content preservation, and overall aesthetics. The results, presented in \tabref{tab:user_study}, demonstrate that our StyleExpert dominates human preference, achieving a Top-1 selection rate of 74.5\%, outperforming all other methods.

\begin{table}[h!]
\centering
\scriptsize
\caption{User study results. Values indicate the Top-1 preference rate (\%) across all methods.}
\vspace{-10pt}
\setlength{\tabcolsep}{2pt} 
\begin{tabular}{@{}lccccccc@{}}
\toprule
      & CSGO & DreamO & OmniGen2 & OmniStyle & QwenEdit & USO & \textbf{StyleExpert} \\ \hline
Top1 & 3.0 & 1.1 & 8.5 & 4.8 & 3.6 & 4.5 & \textbf{74.5} \\ 
\bottomrule
\end{tabular}
\vspace{-12pt}
\label{tab:user_study}
\end{table}

\section{Additional Comparisons}
\label{sec:expanded_comparisons}
\begin{figure*}
    \centering
    \includegraphics[width=0.98\linewidth]{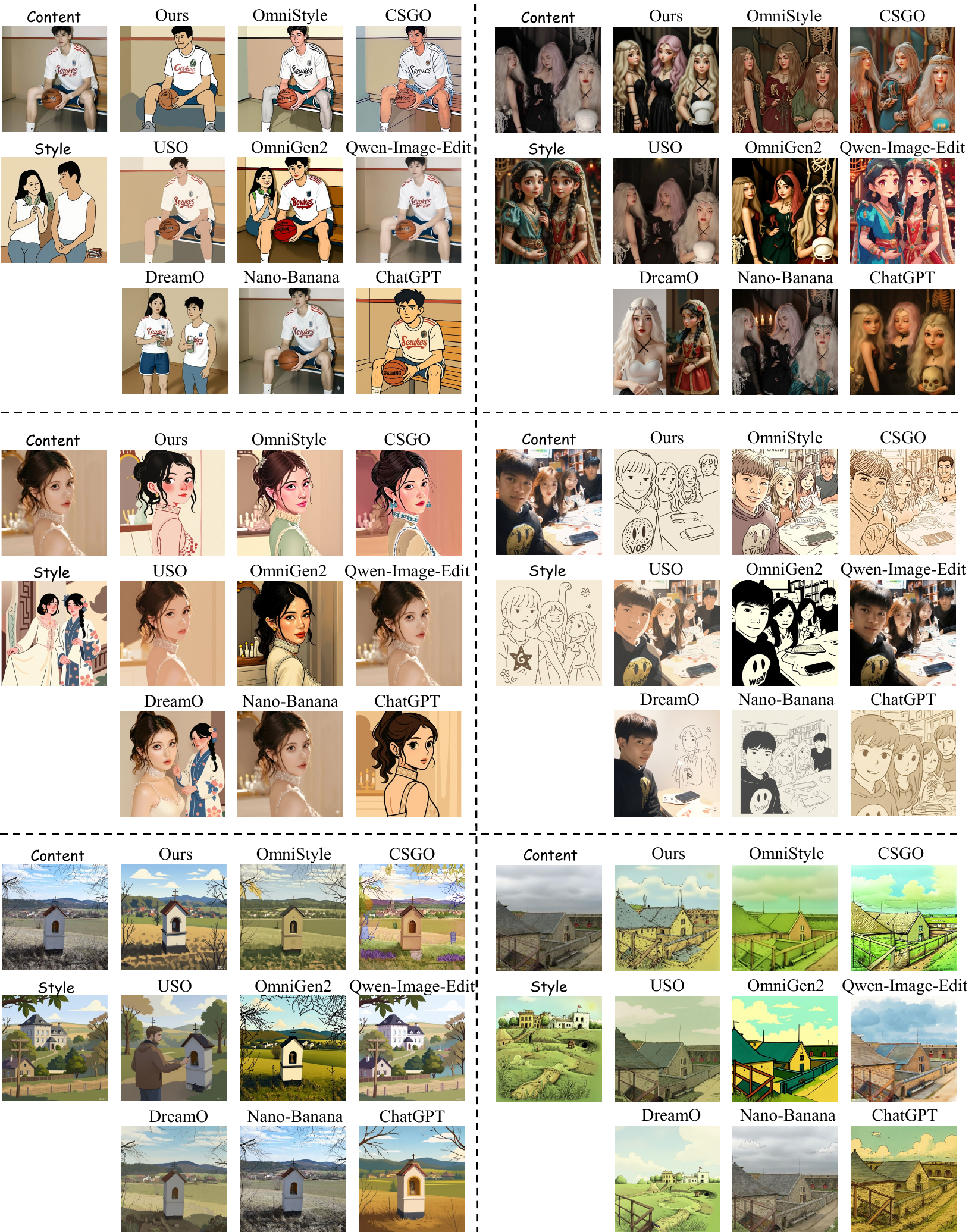}
    \caption{Qualitative comparison of our method against competing approaches under various content and unseen style inputs.}
    \label{fig:more_comparison}
\end{figure*}

\begin{figure*}
    \centering
    \includegraphics[width=0.98\linewidth]{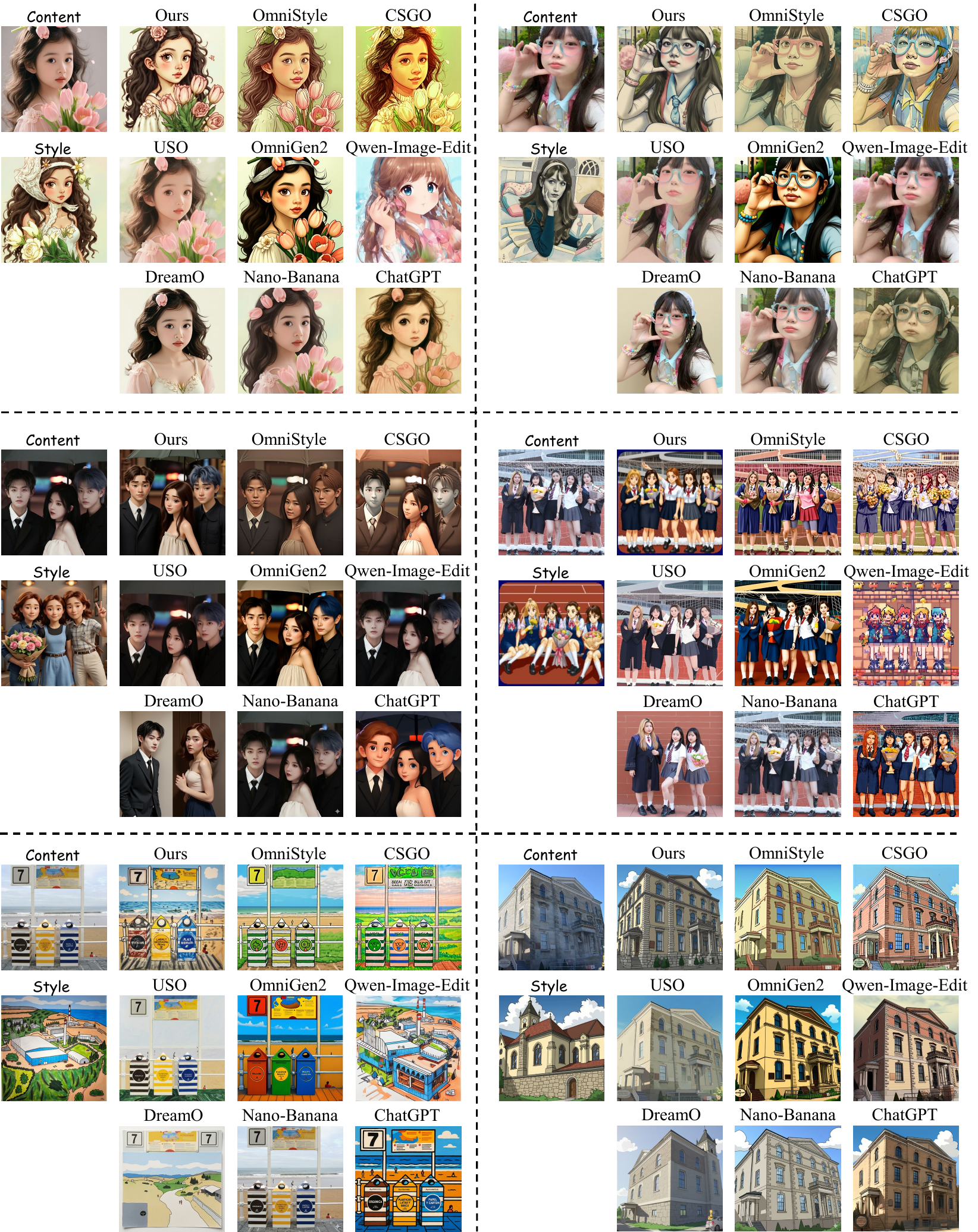}
    \caption{Qualitative comparison of our method against competing approaches under various content and unseen style inputs.}
    \label{fig:more_comparison1}
\end{figure*}

Figures~\ref{fig:more_comparison} and~\ref{fig:more_comparison1} present expanded qualitative comparisons against a broader range of stylization methods.
To provide a more comprehensive analysis, we extend our comparison to include Qwen-Image-Edit \cite{wu2025qwenimagetechnicalreport}, Nano-Banana \cite{nanobanana}, and ChatGPT \cite{gpt}, in addition to the baselines discussed in the main text (OmniStyle \cite{wang2025omnistyle}, CSGO \cite{xing2024csgo}, USO \cite{wu2025uso}, OmniGen2 \cite{wu2025omnigen2}, and DreamO \cite{mou2025dreamo}).
As observed, our method consistently achieves the best overall stylization performance.
In contrast, Qwen-Image-Edit frequently fails to perform stylization, often merely replicating the content or style reference image, or generating irrelevant content (\eg, \figref{fig:more_comparison}, top-right and \figref{fig:more_comparison1}, middle-right).
Similarly, Nano-Banana often fails to capture deep semantic cues from the style reference, resulting in a mere replication of the content image.
While ChatGPT achieves relatively better stylization, it still suffers from insufficient stylization intensity, yielding suboptimal results (\eg, \figref{fig:more_comparison}, top-left and middle-right).

\section{Additional Visual Results}
\label{sec:more_results}
\begin{figure*}
    \centering
    \includegraphics[width=\linewidth]{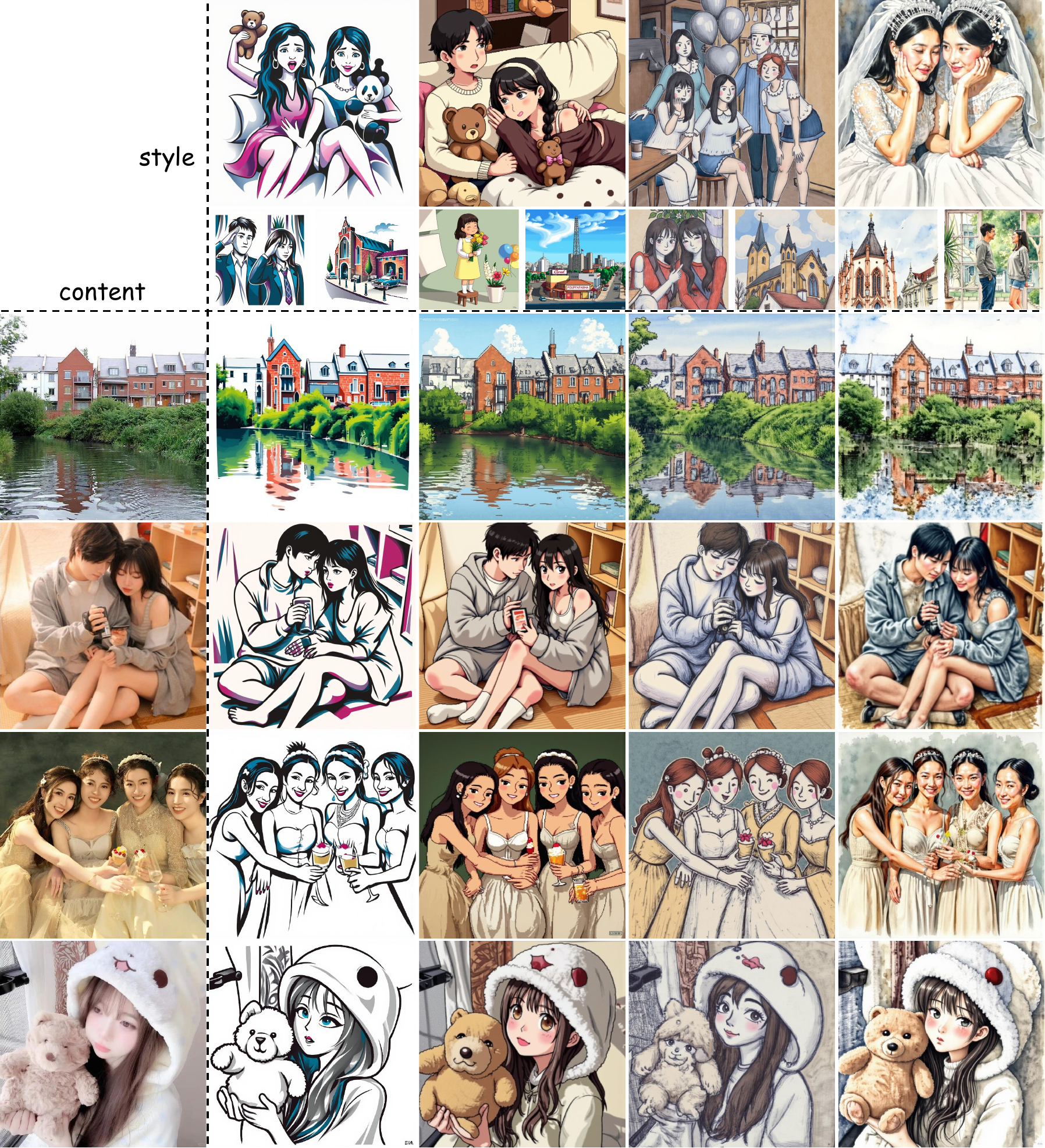}
    \caption{Additional visualization results generated by our method across a diverse range of styles and content subjects.}
    \label{fig:more_visual1}
\end{figure*}

\begin{figure*}
    \centering
    \includegraphics[width=\linewidth]{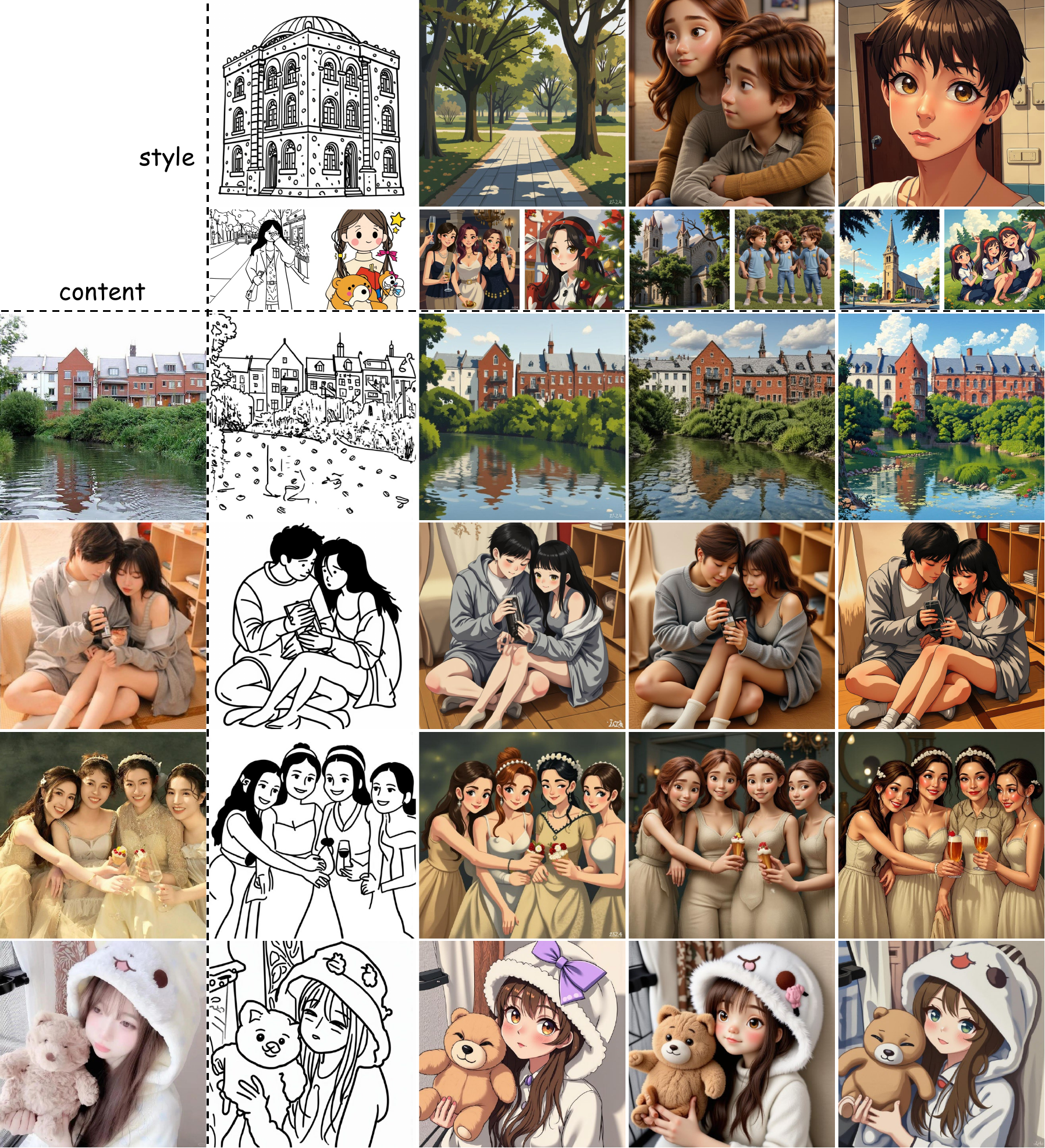}
    \caption{Additional visualization results generated by our method across a diverse range of styles and content subjects.}
    \label{fig:more_visual2}
\end{figure*}

\begin{figure*}
    \centering
    \includegraphics[width=\linewidth]{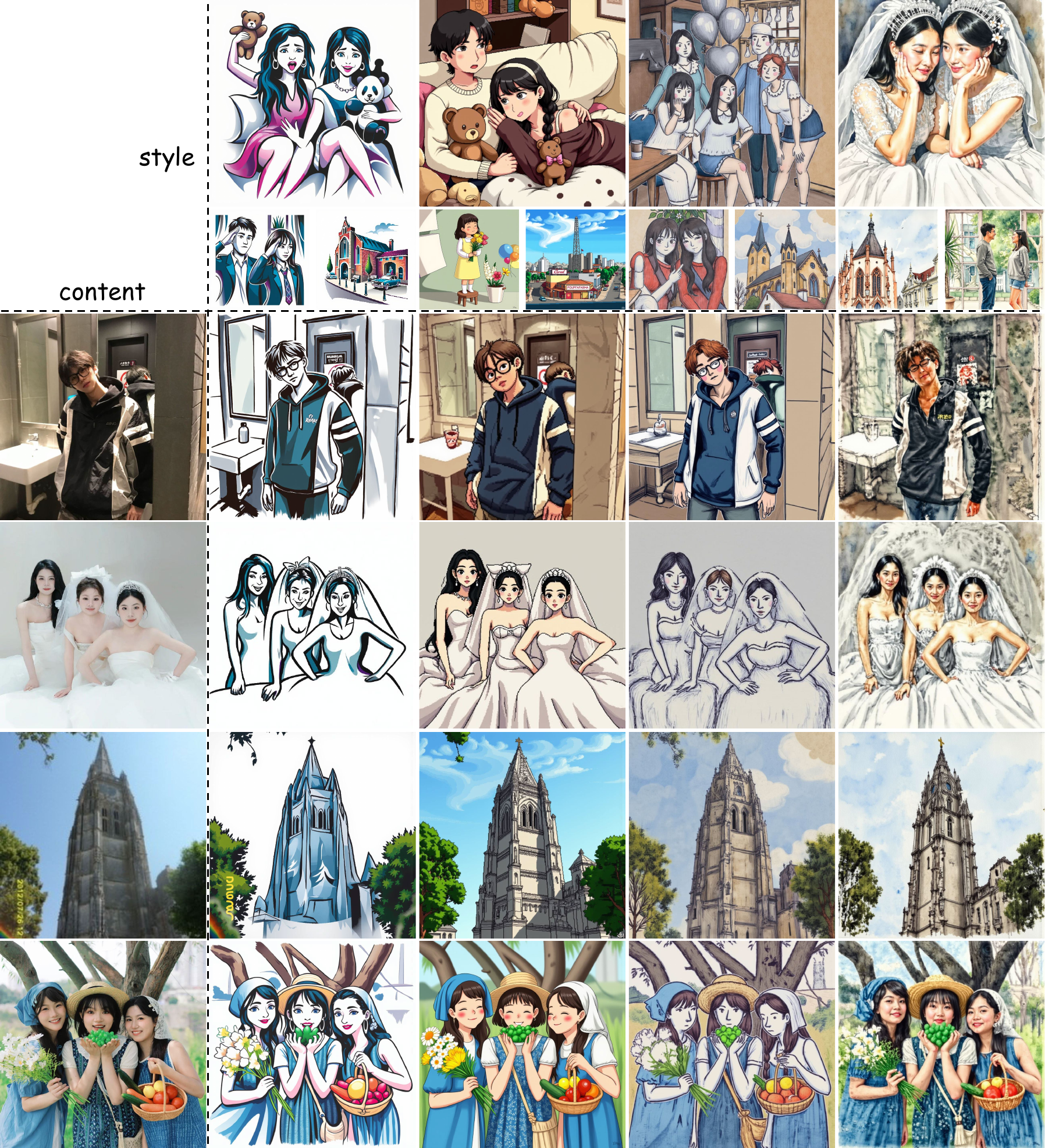}
    \caption{Additional visualization results generated by our method across a diverse range of styles and content subjects.}
    \label{fig:more_visual3}
\end{figure*}

\begin{figure*}
    \centering
    \includegraphics[width=\linewidth]{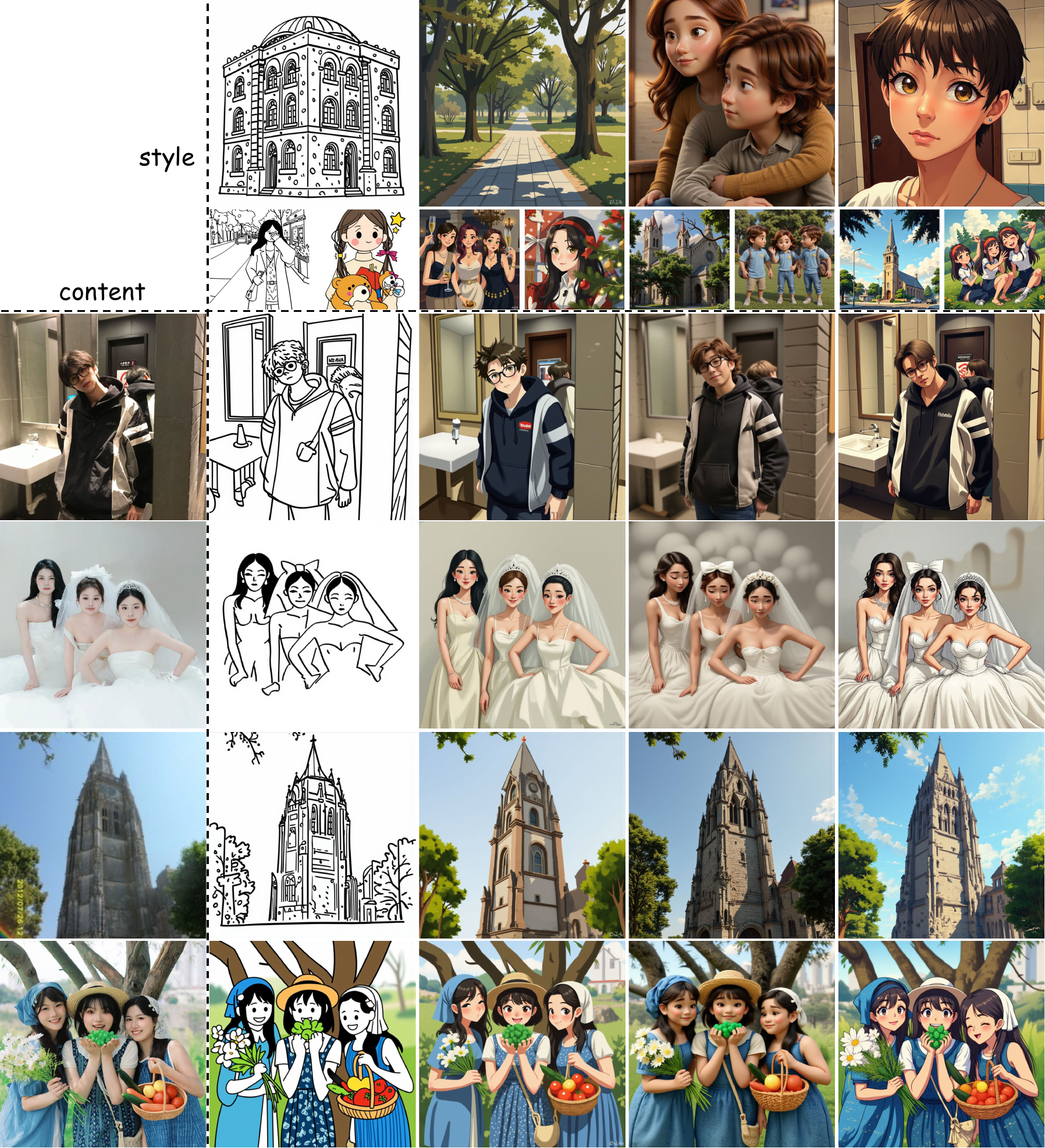}
    \caption{Additional visualization results generated by our method across a diverse range of styles and content subjects.}
    \label{fig:more_visual4}
\end{figure*}

Figures~\ref{fig:more_visual1},~\ref{fig:more_visual2},~\ref{fig:more_visual3}, and \ref{fig:more_visual4} present additional generation results from our method across a diverse range of content and style inputs.
As observed, our method excels not only at traditional color transfer (\eg, \figref{fig:more_visual1}, 1st style column) but also at capturing the overall atmosphere (\eg, \figref{fig:more_visual1}, 4th style column; \figref{fig:more_visual2}, 3rd style column) and transferring intricate textures and lines (\eg, \figref{fig:more_visual1}, 3rd style column; \figref{fig:more_visual2}, 1st style column).
These results fully demonstrate the robustness and reliability of our approach in effectively transferring style attributes across multiple semantic levels.

% ...

\newpage

\begin{figure*}[t]
  \begin{lstlisting}[
    frame=tb,
    language=Python, % 这个内容看起来像Python的dict
    basicstyle=\ttfamily\small, % 使用小号字体，更易读
    caption={The prompt for the Qwen Semantic Score. This prompt instructs the VLM to evaluate whether the stylized output (Style image B, mapped to \texttt{output\_path}) shares the same semantic style (\eg, texture, line quality) as the style reference (Style image A, mapped to \texttt{style\_path}), while ignoring simple color similarities.},
    label={lst:qwen_prompt}
  ]
{
    "role": "user",
    "content": [
        {"type": "text", "text": "You are a visual style analysis expert."},
        {"type": "text", "text": (
              "You will be given two images:\n"
              "1. Style image A\n"
              "2. Style image B\n\n"
              "Your task is to determine whether these two images share the *same artistic style*.\n"
              "By 'same style', we refer to having similar **texture**, **line quality**, and **material or rendering characteristics**, "
              "not merely similar colors, lighting, or atmosphere.\n\n"
              "Answer only 'YES' if both images have the same style, otherwise 'NO'. "
              "Then briefly explain why."
        )},
        {"type": "text", "text": "Style image A:"},
        {"type": "image", "image": style_path},
        {"type": "text", "text": "Style image B:"},
        {"type": "image", "image": output_path},
    ],
}
  \end{lstlisting}
\end{figure*}

\begin{figure*}[t]
  \begin{lstlisting}[
    frame=tb,
    language=Python, % 这个内容看起来像Python的dict
    basicstyle=\ttfamily\small, % 使用小号字体，更易读
    caption={The prompt used to generate initial, content-only captions. This prompt instructs the VLM to describe only the main objects and their spatial relationships, while explicitly forbidding all stylistic, color, or lighting descriptors.},
    label={lst:gen_caption}
  ]
{
    "role": "user",
    "content": [
        {"type": "image", "image": image_path},
        {"type": "text", "text": (
            f"Please generate a caption for this image, "
            f"DO NOT mention style, color, texture, material, lighting, atmosphere, "
            f"or any words like 'red', 'blue', 'green', 'shiny', 'dark', 'bright', "
            f"'realistic', 'light-colored', etc. "
            f"ONLY include main objects and their spatial relationships. "
            f"No more than 50 tokens."
        )},
    ],
}
  \end{lstlisting}
\end{figure*}

\begin{figure*}[t]
  \begin{lstlisting}[
    frame=tb,
    language=Python, % 这个内容看起来像Python的dict
    basicstyle=\ttfamily\small, % 使用小号字体，更易读
    caption={The prompt for the caption rewriting step. This instructs Qwen3 to remove all style, color, and texture descriptors from an input caption (\texttt{caption}).},
    label={lst:rewrite_caption}
  ]
{"role": "system", "content": "You are an assistant that edits text."},
{
    "role": "user", "content": "
    Keep all factual and visual details about objects, people, scenes, and actions.
    Remove all references to style, color, texture, material, lighting, or atmosphere, such as:
    red, blue, green, shiny, dark, bright, realistic, light-colored, etc.
    Do not add new information.
    Output only the cleaned description.
    
    Based on the rules above, rewrite the following description in English:
    {caption}
    "
}
  \end{lstlisting}
\end{figure*}

\begin{figure*}[t]
  \begin{lstlisting}[
    frame=tb,
    language=Python, % 这个内容看起来像Python的dict
    basicstyle=\ttfamily\small, % 使用小S号字体，更易读
    caption={The prompt for the content-matching filter. This instructs the VLM to verify if the stylized image (\texttt{img\_path}) strictly matches the clean content caption (\texttt{caption}).},
    label={lst:filter1}
  ]
{
{
    "role": "user",
    "content": [
        {"type": "text", "text": "Here is a stylized image:"},
        {"type": "image", "image": img_path},
        {"type": "text", "text": (
            f"Check if the content of this image strictly matches the caption: "
            f" '{caption}'."
            f"Ignore color, material, and things related to style in caption, "
            f"answer only 'YES' if it strictly matches, otherwise 'NO'. "
            f"Then briefly explain why."
        )},
    ],
}
  \end{lstlisting}
\end{figure*}

\begin{figure*}[t]
  \begin{lstlisting}[
    frame=tb,
    language=Python, % 这个内容看起来像Python的dict
    basicstyle=\ttfamily\small, % 使用小号字体，更易读
    caption={The prompt for the person-consistency filter. This instructs the VLM to verify that the stylized image (\texttt{img\_path}) and the source image (\texttt{src\_path}) contain the same number of people and that their genders strictly match.},
    label={lst:filter2}
  ]
{
    "role": "user",
    "content": [
        {"type": "text", "text": "This is image 1:"},
        {"type": "image", "image": img_path},
        {"type": "text", "text": "This is image 2:"},
        {"type": "image", "image": src_path},
        {"type": "text", "text": (
            f"Check if the two images"
            f"1. Contains the same NUMBER of objects (\eg, people). Variations that reflect intentional youthful stylization should be recognized as the same person."
            f"2. GENDER OF EACH PERSON strictly matches"
            f"YOU MUST IGNORE ANY OTHER DETAIL CHANGE AND FOCUS ONLY ON NUMBER & EACH PERSON'S GENDER."
            f"Answer only 'YES' if the number & EACH PERSON's gender met, otherwise 'NO'."
            f"Then briefly explain why."
        )}
    ],
}
  \end{lstlisting}
\end{figure*}

\end{document}